\documentclass{article}

     \usepackage[final,nonatbib]{neurips_2020}

\usepackage[utf8]{inputenc} 
\usepackage[T1]{fontenc}    
\usepackage{hyperref}       
\usepackage{url}            
\usepackage{booktabs}       
\usepackage{amsfonts}       
\usepackage{nicefrac}       
\usepackage{microtype}      
\usepackage{subcaption}

\usepackage{amsmath}
\usepackage{amssymb}
\usepackage[capitalise,noabbrev]{cleveref}
\usepackage{algorithm}
\usepackage{algorithmic}
\usepackage{verbatim}
\usepackage{graphicx}
\usepackage{xcolor}
\usepackage{glossaries}
\glsdisablehyper
\usepackage{subcaption}
\usepackage{todonotes}
\usepackage{tablefootnote}
\usepackage[frozencache]{minted}

\title{Sample-Efficient Optimization in the Latent Space of\\Deep Generative Models via Weighted Retraining}

\author{%
Austin Tripp\thanks{equal contribution}\\
University of Cambridge\\
\texttt{ajt212@cam.ac.uk}
\And
Erik Daxberger\footnotemark[1]\\
University of Cambridge\\
Max Planck Institute for\\
Intelligent Systems, Tübingen\\
\texttt{ead54@cam.ac.uk}
\And
Jos\'e Miguel Hern\'andez-Lobato\\
University of Cambridge\\
Alan Turing Institute\\
Microsoft Research\\
\texttt{jmh233@cam.ac.uk}

}

\newcommand{\D}{\mathcal{D}}
\newcommand{\X}{\mathcal{X}}
\newcommand{\Z}{\mathcal{Z}}

\newcommand{\x}{\mathbf{x}}
\newcommand{\z}{\mathbf{z}}

\newacronym{dgm}{DGM}{deep generative model}
\newacronym{lmo}{LSO}{latent space optimization}
\newacronym{vae}{VAE}{variational autoencoder}
\newacronym{gan}{GAN}{generative adversarial network}
\newacronym{rl}{RL}{reinforcement learning}
\newacronym{bo}{BO}{Bayesian optimization}
\newacronym{sgp}{SGP}{sparse Gaussian process}

\newcommand{\codelink}{\url{https://github.com/cambridge-mlg/weighted-retraining}
}
\newcommand{\bestchemscore}{27.84}

\begin{document}

\maketitle

\begin{abstract}
Many important problems in science and engineering, such as drug design, involve
optimizing an expensive black-box objective function over a complex, high-dimensional, and structured input space.
Although machine learning techniques have shown promise in solving such problems, existing approaches substantially lack sample efficiency.
We introduce an improved method for efficient black-box optimization,
which performs the optimization in the low-dimensional, continuous latent manifold learned by a deep generative model.
In contrast to previous approaches, we actively steer the generative model to maintain a latent manifold that is highly useful for efficiently optimizing the objective.
We achieve this by periodically \emph{retraining} the generative model on the data points queried along the optimization trajectory, as well as \emph{weighting} those data points according to their objective function value.
This weighted retraining can be easily implemented on top of existing methods, 
and is empirically shown to significantly improve their efficiency and performance on synthetic and real-world optimization problems.
\end{abstract}
\section{Introduction}
\label{sec:introduction}
Many important problems in science and engineering can be formulated as optimizing an objective function over an input space.
Solving such problems becomes particularly challenging when 1) the input space is high-dimensional and/or \emph{structured} (i.e.\@ discrete spaces, or non-Euclidean spaces such as graphs, sequences, and sets)
and 2) the objective function is expensive to evaluate.
Unfortunately, many real-world problems of practical interest have these characteristics.
A notable example is drug design, which has a graph-structured input space, and is
evaluated using expensive wet-lab experiments or time-consuming simulations.
Recently, machine learning has shown promising results in many problems that can be framed as optimization, such as 
conditional image \cite{van_den_oord_conditional_2016,nguyen_plug_2017} and text \cite{otter_survey_2020} generation,
molecular and materials design \cite{elton_deep_2019,sanchez-lengeling_inverse_2018},
and neural architecture search \cite{elsken_neural_2019}.
Despite these successes, using machine learning on structured input spaces and with limited data
is still an open research area, making the use of machine learning
infeasible for many practical applications.

One promising approach which tackles \emph{both} challenges is a two-stage procedure that has emerged over the past few years,
which we will refer to as \emph{\gls{lmo}} \cite{gomez2018,kusner_grammar_2017,lu2018structured,luo2018neural,nguyen_synthesizing_2016}.
In the first stage, a (deep) generative model is trained to map tensors in a low-dimensional continuous space
onto the data manifold in input space,
effectively constructing a low-dimensional and continuous analog of the optimization problem.
In the second stage, the objective function is optimized over this learned latent space using a surrogate model.
Despite many successful applications in a variety of fields including chemical design \cite{gomez2018,jin_junction_2019,kusner_grammar_2017,dai_syntax-directed_2018} and automatic machine learning \cite{lu2018structured,luo2018neural},
\gls{lmo} is primarily applied in a \textit{post-hoc} manner using a pre-trained, general purpose generative model
rather than one trained specifically for the explicit purpose of downstream optimization.
Put differently, the training of the generative model is effectively \emph{decoupled} from the optimization task.

\begin{figure}[tb]
    \centering
    \begin{subfigure}[c]{0.245\textwidth}
        \centering
        \includegraphics[width=0.9\textwidth]{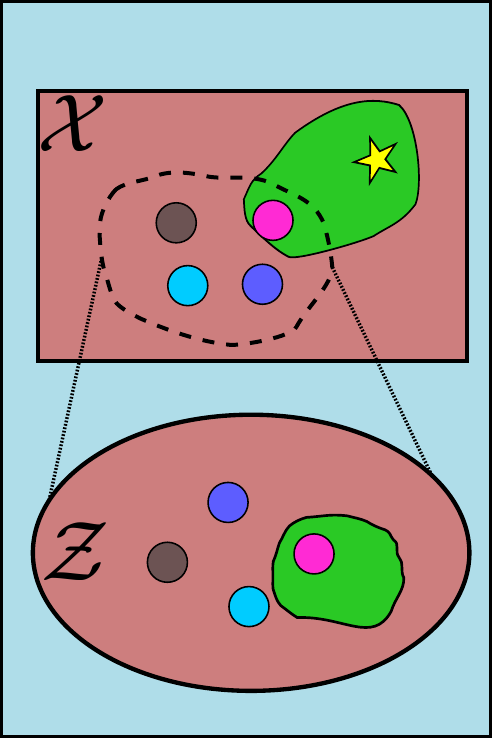}
        \vspace{-5mm}
        \includegraphics[width=0.95\textwidth]{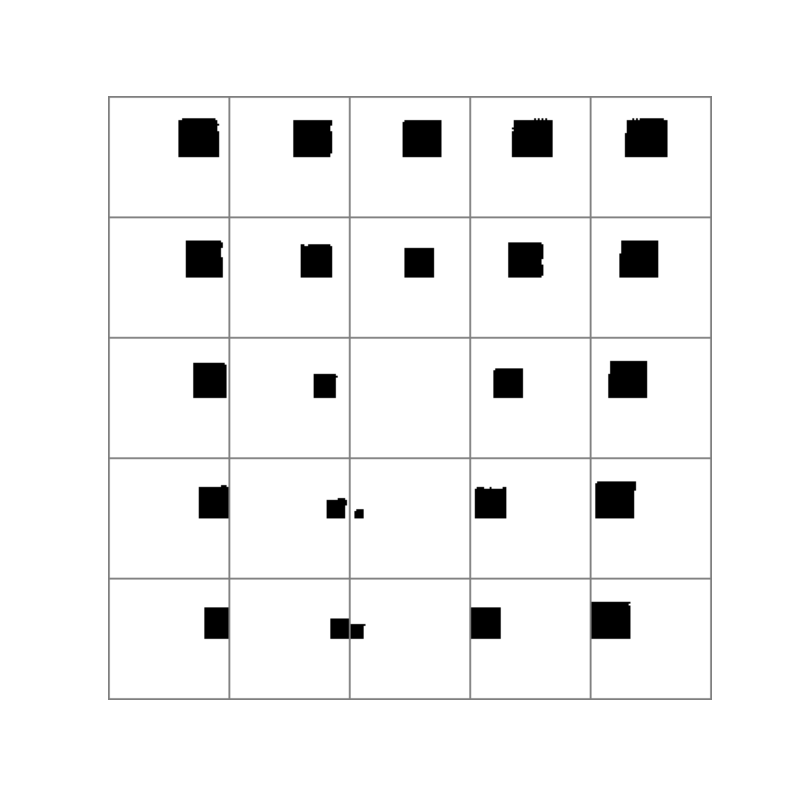}
        \subcaption{Starting Point}
        \label{subfig:schematic-a}
    \end{subfigure}
    \begin{subfigure}[c]{0.245\textwidth}
        \centering
        \includegraphics[width=0.9\textwidth]{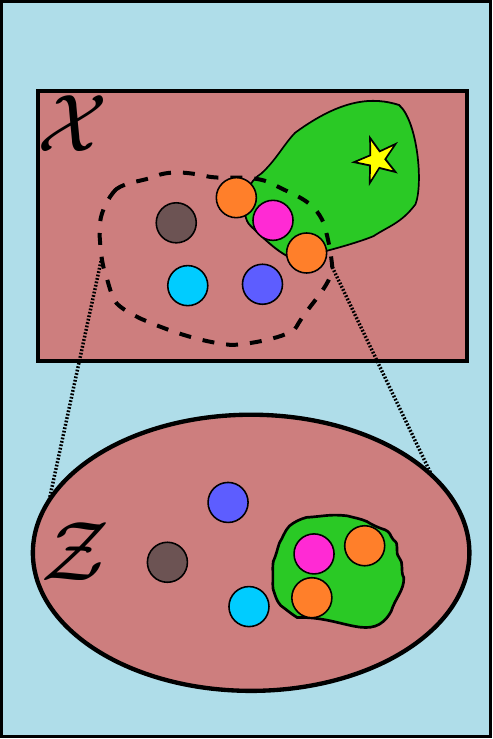}
        \vspace{-5mm}
        \includegraphics[width=0.95\textwidth]{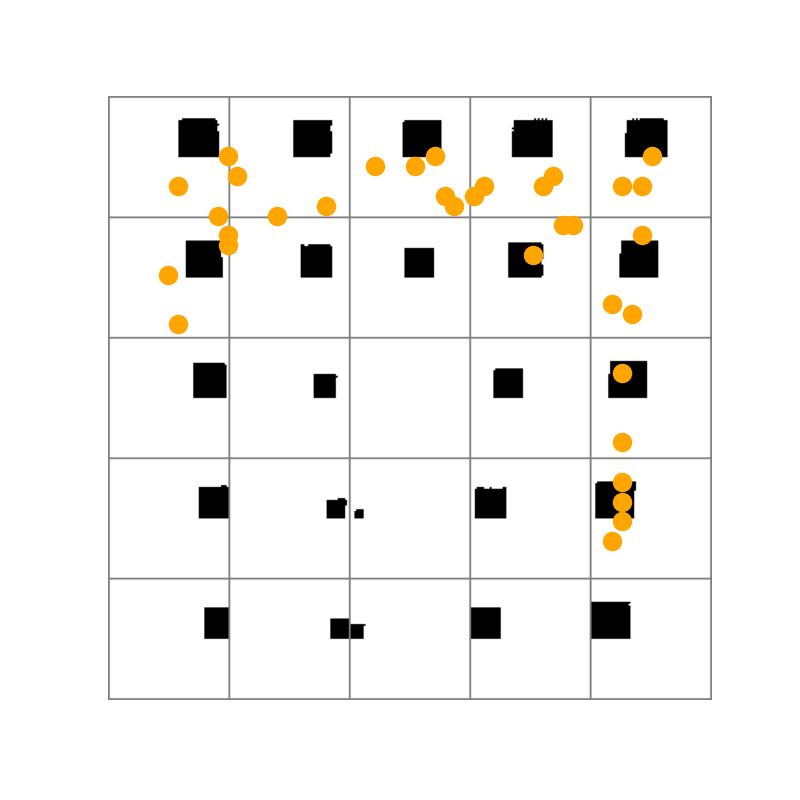}
        \subcaption{Standard \gls{lmo}}
        \label{subfig:schematic-b}
    \end{subfigure}
    \begin{subfigure}[c]{0.49\textwidth}
        \centering
        \includegraphics[width=0.9\textwidth]{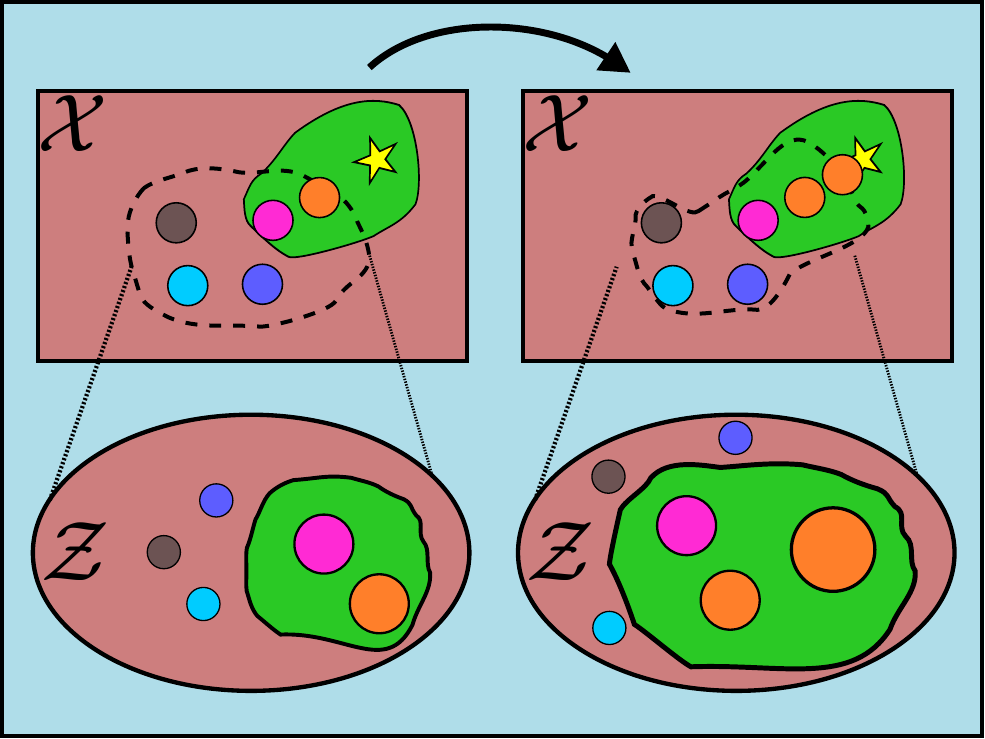}
        \vspace{-5mm}
        \includegraphics[width=0.475\textwidth]{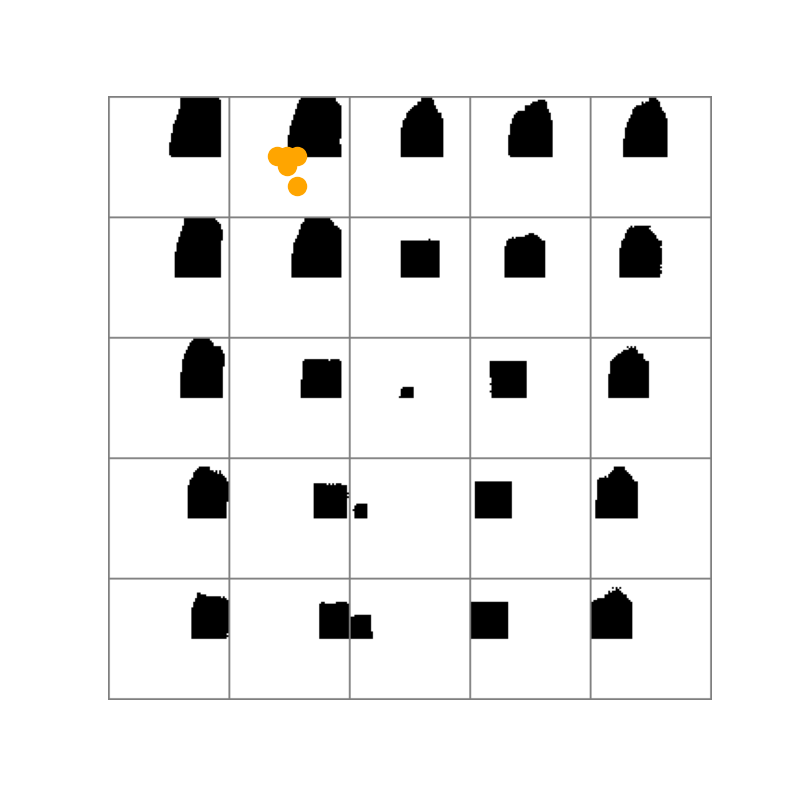}
        \includegraphics[width=0.475\textwidth]{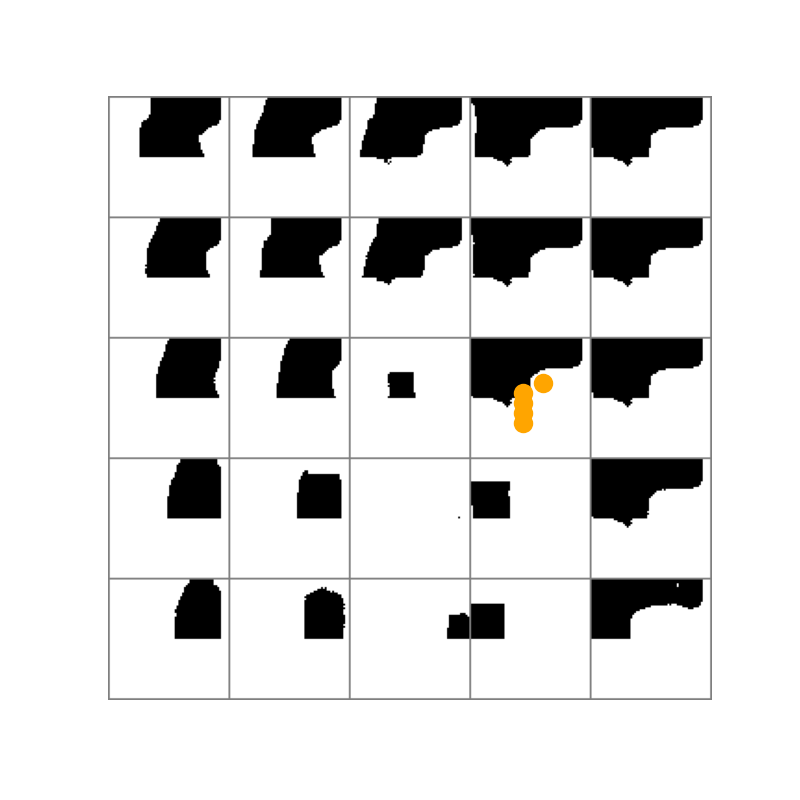}
        \subcaption{\Gls{lmo} with Weighted Retraining}
        \label{subfig:schematic-c}
    \end{subfigure}
    \caption{
    Schematic illustrating \gls{lmo} with and without weighted retraining.
    The cartoon illustrates the input/latent space  of the generative model (\textbf{top}).
    The latent manifold from \cref{subsec:expt-wr}'s 2D shape area maximization task is shown
    for comparison (\textbf{bottom}).
    Each image in the manifold shows the result of decoding a latent point on a uniform square grid in a 2D latent space;
    images are centered on the original grid points.
    Red/green regions correspond to points with low/high objective function values respectively.
    The yellow star is the global optimum in $\X$.
    Coloured circles are data points; their radius represents their weight.
    The dashed line surrounds the region of $\X$ modelled by $g$ (i.e.\@ $g(\Z)$, the image of $\Z$).
    \textbf{(a)} The status of the generative model $g$ at the start of optimization.
    \textbf{(b)} The result of standard \gls{lmo} with $g$ fixed, which queries the points in orange.
    It is only able to find points close to the training data used to learn $\Z$, resulting in slow and incomplete exploration of $\mathcal{X}$.
    \textbf{(c)} The result midway (left) and at the end (right) of \gls{lmo} with our proposed approach,
    which weights data points according to their objective function value and retrains $g$ to incorporate newly queried data.
    This continually adjusts $\Z$ to focus on modelling the most promising regions of $\X$,
    speeding up the optimization and allowing for substantial extrapolation beyond the initial training data.
    }
    \label{fig:schematic}
\end{figure}

In this work, we identify and examine two types of decoupling in \gls{lmo}.
We argue that they make optimization unnecessarily difficult and fundamentally prevent \gls{lmo} from 
finding solutions that lie far from the training data.
Motivated by this, we propose \emph{weighting of the data distribution} and \emph{periodic retraining of the generative model} to effectively resolve this decoupling.
We argue that these two modifications are highly complementary,
fundamentally transforming \gls{lmo} from a local optimizer into an efficient global optimizer
capable of recursive self-improvement.
Our contributions are:
\begin{enumerate}
    \item We identify and describe two critical failure modes of previous \gls{lmo}-based methods which severely limit their efficiency and performance, and thus practical applicability (\cref{sec:limitations}).
    \item We propose to combine dataset weighting with periodic retraining of the generative model used within \gls{lmo} as an effective way to directly address the issued identified (\cref{sec:main}).
    \item We empirically demonstrate that weighted retraining significantly benefits \gls{lmo} across a variety of application domains and generative models,
    achieving substantial improvements over state-of-the-art methods on a widely-used chemical design benchmark (\cref{sec:experiments}).
\end{enumerate}

\section{Problem Statement and Background}
\label{sec:background}
\textbf{Sample-Efficient Black Box Optimization.}
Let $\X$ be an \textit{input space}, and let $f: \X\mapsto\mathbb{R}$ be an \textit{objective function}.
In particular, we focus on cases where 1) the input space $\X$ is \emph{high-dimensional} (i.e.\@ $100+$ effective dimensions) and \emph{structured} (e.g. graphs, sequences or sets),
and 2) the objective function $f(\x)$ is \emph{black-box} (i.e.\@ no known analytic form or derivative information available) and is \emph{expensive} to evaluate (e.g.\@ in terms of time or energy cost).
The \emph{sample-efficient optimization} problem seeks to optimize $f$ over $\X$,
evaluating $f$ as few times as possible, producing in the process a sequence of evaluated points
$\mathcal{D}_M\equiv \{\x_i, f(\x_i)\}_{i=1}^M$ with $M$ function evaluations.

\textbf{Model-Based Optimization}
seeks to approximate $f$ by constructing an \textit{objective model} $h_{\X}:\X\mapsto\mathbb{R}$.
$h_{\X}$ is then optimized as a surrogate for $f$.
Although this is a widely used optimization technique,
it can be difficult to apply in high-dimensional spaces (which may have many local minima)
or in structured spaces (where any kind of discreteness precludes the use of gradient-based optimizers).

\textbf{Latent Space Optimization (LSO)}
is a technique wherein model-based optimization is performed in the \textit{latent space} $\Z$
of a 
\textit{generative model} $g: \Z\mapsto\X$ that maps from $\Z$ to input space $\X$.
To this end, a \textit{latent objective model} $h: \Z\mapsto\mathbb{R}$ is constructed to approximate $f$ at the output of $g$, i.e.\@ such that $f(g(\z))\approx h(\z),\ \forall \z\in\Z$.
If $\Z$ is chosen to be a low-dimensional, continuous space such as $\mathbb{R}^n$,
the aforementioned difficulties with model-based optimization can be avoided,
\emph{effectively turning a discrete optimization problem into a continuous one}.
To realize \gls{lmo}, $g$ can be chosen to be a state-of-the-art \gls{dgm},
such as a \gls{vae} \cite{kingma2013auto,rezende2014stochastic} or a \gls{gan} \cite{goodfellow2014generative},
which have been shown to be capable of learning vector representations of many types of high-dimensional,
structured data \cite{bowman_generating_2016,de_cao_molgan_2018,simonovsky_graphvae_2018,wang_graphgan_2018}.
Furthermore, $h$ can be chosen to be a flexible probabilistic model such as a Gaussian process \cite{williams2006gaussian},
allowing sample-efficient \emph{Bayesian optimization} to be performed \cite{brochu2010tutorial,shahriari2015taking}.
$h$ can be trained by using an approximate inverse to $g$, $q:\X\mapsto\Z$, to find a corresponding latent point $\z_i$ for each data point $\x_i$.

\section{Failure Modes of Latent Space Optimization}
\label{sec:limitations}
To understand the shortcomings of \gls{lmo},
it is necessary to first examine in detail the role of the generative model, which is usually a \gls{dgm}.
State-of-the-art \glspl{dgm} such as \glspl{vae} and \glspl{gan}
are trained with a prior $p(\z)$ over the latent space $\Z$.
This means that although the resulting function $g:\Z\mapsto\X$ is defined over the entire latent space $\Z$,
it is effectively only trained on points in regions of $\Z$ with high probability under $p$.
Importantly, even if $\Z$ is an unbounded space with infinite volume such as $\mathbb{R}^n$,
because $p$ has finite volume, there must exist a \emph{finite} subset $\Z'\subset\Z$ that contains virtually all the probability mass of $p$.
We call $\Z'$ the \emph{feasible region} of $\Z$.
Although in principle optimization can be performed over all of $\Z$,
it has been widely observed that optimizing outside of the feasible region tends to give poor results,
yielding samples that are low-quality, or even invalid (e.g.\@ invalid molecular strings, non-grammatical sentences);
therefore all \gls{lmo} methods known to us employ some sort of measure to restrict the optimization to near or within the feasible region
\cite{gomez2018,kusner_grammar_2017,nguyen_synthesizing_2016,griffiths_constrained_2020,white_sampling_2016,mahmood_cold_2019,daxberger2019bayesian}.
This means that \gls{lmo} should be treated as a \emph{bounded} optimization problem,
whose feasible region is determined by $p$.

Informally, the training objective of $g$ encourages points sampled from within the feasible region to match the data distribution that $g$
was trained on, effectively ``filling'' the feasible region with points similar to the dataset,
such that a point's relative volume is roughly proportional to its frequency in the training data.
For many optimization problems, most of the training data for the \gls{dgm} is low-scoring
(i.e.\@ highly sub-optimal objective function values),
thereby causing most of the feasible region to contain low-scoring points.
Not only does this make the optimization problem more difficult to solve (like finding the proverbial ``needle in a haystack''),
but actually leaves insufficient space in the feasible region for a large number of novel, high-scoring points that lie outside the training distribution to be modelled by the \gls{dgm}.
Therefore, even a perfect optimization algorithm with unlimited evaluations of the objective function
might be unable to find a novel point
that is substantially better than the best point in the original dataset,
simply because such a point may not exist in the feasible region.

This pathological behaviour is conceptually illustrated in \cref{subfig:schematic-b},
where \gls{lmo} is unable to find or even approach the global optimum that lies far from the training data.
We propose that \gls{lmo}'s performance is severely limited by two concrete problems in its setup.
The first problem is that the generative model's training objective (to learn a latent space that captures the data distribution as closely as possible),
does not necessarily match the true objective (to learn a latent space that is amenable to efficient optimization of the objective function).
Put in terms of the cartoon in \cref{subfig:schematic-b}, the feasible region that is learned, which uniformly and evenly surrounds the data points,
is not the feasible region that would be useful for optimization,
which would model more of the green region at the expense of the red region.
This is also seen in the 2D shape area maximization task
in \cref{subfig:schematic-b},
where the latent manifold contains only low-area shapes that the model was trained on,
and nothing close to the all-black global optimum.
The second problem is that information on new points acquired during the iterative optimization procedure
is not propagated to the generative model,
where it could potentially help to refine and expand the coverage of the feasible region,
uncovering new promising regions that an optimization algorithm can exploit.
In terms of \cref{subfig:schematic-b},
the new data is not used to shift the feasible region toward the green region, despite the optimization process indicating that this is a very promising region of $\X$ for optimization.
Luckily, we believe that neither of these two problems is inherent to \gls{lmo}, and now pose a framework that directly addresses them.

\section{Latent Space Optimization with Weighted Retraining}
\label{sec:main}
\subsection{Training a Generative Model with a Weighted Training Objective}
\label{subsec:weighting}
While it is unclear in general how to design a generative model that is maximally amenable to \gls{lmo},
the argument presented in \cref{sec:limitations} suggests that it would at least be beneficial to dedicate
a higher fraction of the feasible region to modelling high-scoring points.
One obvious but inadequate method of achieving this is to simply discard all low-scoring points from the dataset used to train the \gls{dgm}, e.g.\@ by keeping only the top 10\% of the data set (in terms of score).
While this strategy could be feasible if data is plentiful, when data is scarce this option may not be viable
because state-of-the-art neural networks need a large amount of training data to avoid overfitting.
This issue can be resolved by not viewing inclusion in the dataset as a binary choice,
but instead as a \emph{continuum} that can be realized by \emph{weighting} the data points unevenly.
If the generative model is trained on a distribution that systematically places more probability mass on high-scoring points
and less mass on slow scoring points,
the distribution-matching term in the \gls{dgm}'s training objective will incentivize a larger fraction of the feasible
region's volume to be used to model high-scoring points,
while simultaneously using all known data points to learn useful representations and avoid overfitting.

A simple way to achieve this weighting is to assign an explicit weight $w_i$ to each data point, such that $\sum_i w_i=1$.
As the training objective of common \glspl{dgm} involves the expected value of a loss
function $\mathcal{L}$ with respect to the data distribution,%
\footnote{For a \gls{vae}, $\mathcal{L}$ is the per-datapoint ELBO \cite{kingma2013auto}, while for a \gls{gan}, $\mathcal{L}$ is the discriminator score \cite{goodfellow2014generative}.}
weighted training can be implemented by simply replacing the empirical mean over the training data with a \emph{weighted} empirical mean:
i.e.\@ $\sum_{\x_i \in \D} w_i\mathcal{L}(\x_i)$ instead of $\sum_{\x_i \in \D} \frac{1}{N} \mathcal{L}(\x_i)$.
In practice, mini-batch stochastic gradient descent is used to optimize this objective
to avoid summing over all data points.
Unbiased mini-batches can be constructed by sampling each data point $\x_i$ with probability $w_i$
with replacement to construct each batch (see \cref{subsec:weighted-mini-batching} for more details).

\begin{figure}[ht]
    \hspace{-4mm}
    \includegraphics{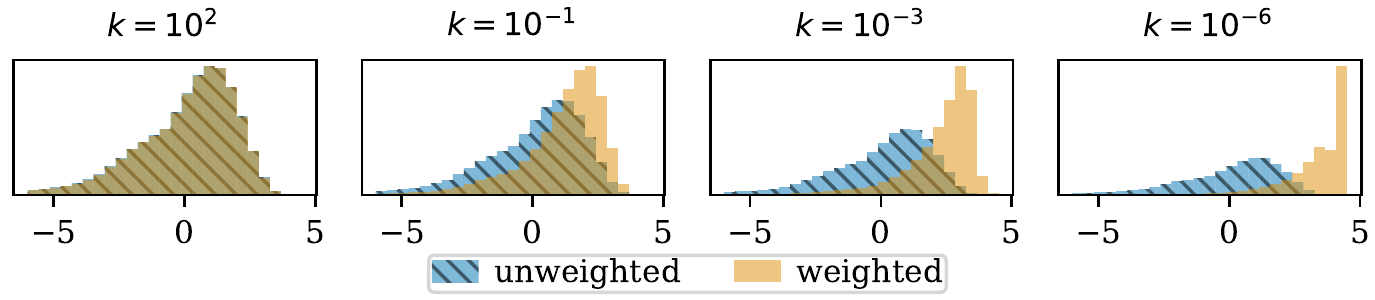}
    \hspace{-4.5mm}
    \caption{
    Histogram of objective function values for the ZINC dataset (see \cref{sec:experiments})
    with uniform weighting (in blue) as well as rank weighting from \cref{eq:weighting_function} for different $k$ values (in orange).
    Large $k$ approaches uniform weighting, while small $k$ places most weight on high-scoring points.
    }
    \label{fig:weighting}
\end{figure}

We offer no universal rules for setting weights, except that all weights $w_i$ should be restricted to strictly positive values,
because a negative weight would incentivize the model to perform poorly,
and a weight of zero is equivalent to discarding a point.
This aside, there are many reasonable ways to choose the weights such that high-scoring points are weighted more,
and low-scoring points are weighted less.
In this work, we decide to use a rank-based weight function,
\begin{equation}
\label{eq:weighting_function}
    w(\x; \D, k) \propto \frac{1}{kN + \text{rank}_{f,\D}(\x)}, \quad \text{rank}_{f,\D}(\x) = \left|\left\{\x_i : f(\x_i) > f(\x),\ \x_i \in \D \right\}\right|\ ,
\end{equation}
which assigns a weight roughly proportional to the reciprocal (zero-based) rank of each data point.
We chose \cref{eq:weighting_function} because it yields weights which are always positive,
resilient to outliers, and has stable behaviour over a range of dataset sizes (this is explained further in \cref{subsec:rank-weights}).
Furthermore, as shown in \cref{fig:weighting}, it admits a single tunable hyperparameter $k$ which continuously controls the degree of weighting,
where $k=\infty$ corresponds to uniform weighting, i.e. $w_i = \frac{1}{N}, \forall i$, while $k=0$ places \emph{all} mass on only the single point with the highest objective function value.

\subsection{Periodic Retraining to Update the Latent Space}
\label{subsec:retraining}
To allow the latent manifold to adapt to new information,
we propose a conceptually simple solution: \emph{periodically retraining the generative model} during the optimization procedure.
In practice, this could be done by training a new model from scratch,
or by fine-tuning the previously trained model on the novel data.
However, as is often pointed out in the active learning literature,
the effect of adding a few additional points to a large dataset is rather negligible,
and thus it is unlikely that the generative model will change significantly if retrained on
this augmented dataset \cite{sener_active_2018}.
While one could also retrain on \emph{only} the new data,
this might lead to the well-known phenomenon of catastrophic forgetting \cite{mccloskey_catastrophic_1989}.

A key observation we make is that \emph{the data weighting outlined in \cref{subsec:weighting} actually resolves this problem}.
Specifically, if the new points queried are high-scoring,
then a suitable weighting scheme (such as \cref{eq:weighting_function}) will assign a large weight to them,
while simultaneously decreasing the weights of many of the original data points,
meaning that a small number of new points can have a disproportionate impact on the training distribution.
If the generative model is then retrained using this distribution,
it can be expected to change significantly to incorporate these new points into the the latent space in order to minimize the weighted loss.
By contrast, if the new points queried are low-scoring,
then the distribution will change negligibly, and the generative model will not significantly update, thereby avoiding adding new low-scoring points into the feasible region.
\subsection{Weighted Retraining Combined}
\label{subsec:weighted-retraining}
When put together, \emph{data weighting} and \emph{periodic retraining} complement each other elegantly,
transforming the generative model from a passive decoding function into an active participant
in the optimization process,
whose role is to ensure that the latent manifold is constantly
occupied by the most updated and relevant points for optimization.
Their combined effect is visualized conceptually in \cref{subfig:schematic-c}.
In the first iteration, weighted training creates a latent space with more high scoring points,
causing the feasible region to extend farther into the green region at the expense of the red region.
This allows a better orange point to be chosen relative to \cref{subfig:schematic-b}.
In the second iteration in \cref{subfig:schematic-c},
weighted training with the orange point 
incorporates even more high-scoring points into the latent space,
allowing an even better point to be found.
Qualitatively similar results can be seen in the 2D shape area maximization task,
where weighted retraining introduces points with very high areas
into the latent space compared to \cref{subfig:schematic-a}
(details for this experiment are given in \cref{sec:experiments}).

\begin{algorithm}[tb]
\caption{Latent Space Optimization {\color{blue} with Weighted Retraining} (changes highlighted in {\color{blue} blue})}
\label{alg:algorithm}
\begin{algorithmic}[1]
    \STATE {\bfseries Input:} Data $\D = \{(\x_i, f(\x_i))\}_{i=1}^N$, query budget $M$, objective function $f(\x)$, latent objective model $h(\z)$,
    generative/inverse model $g(\z)$/$q(\x)$, {\color{blue} retrain frequency $r$, weighting function $w(\x)$}
    \vspace{1mm}
    {\color{blue}\FOR{$1, \ldots, M$\textcolor{blue}{$/r$}}
        {\color{black} \STATE Train generative model $g$ and inverse model $q$ on data $\D$ {\color{blue} weighted by $\mathcal{W} = \{w(\x)\}_{\x \in \D}$}%
        \vspace{1mm}
    	\FOR{$1, \ldots, $\textcolor{blue}{$r$}}
            \STATE Compute latent variables $\mathcal{Z} = \{\z = q(\x)\}_{\x \in \D}$
            \STATE Fit objective model $h$ to $\mathcal{Z}$ and $\D$,
            and optimize $h$
            to obtain new latent query point $\tilde{\z}$
    	    \STATE Obtain corresponding input $\tilde{\x} = g(\tilde{\z})$, evaluate $f(\tilde{\x})$ and set $\D \gets \D \cup \{(\tilde{\x}, f(\tilde{\x}))\}$
        \ENDFOR}
    \ENDFOR}
    \vspace{1mm}
	\STATE {\bfseries Output:} Augmented dataset $\D$
\end{algorithmic} 
\end{algorithm}

In the remainder of this paper, we refer to the combination of these techniques as
\emph{weighted retraining} for brevity; see \cref{alg:algorithm} for pseudocode.
We highlight that this algorithm is straightforward to implement in most models,
with brief examples given in \cref{subsec:weighted-training-implementation}.
Computationally, the overhead of the weighting is minimal, and the cost of the retraining
can be reduced by fine-tuning an existing model on the weighted dataset instead of retraining it from scratch.
Although this may still be prohibitively expensive for some applications,
we stress that in many scenarios the cost of training a model is insignificant compared to even a single evaluation of the objective function (e.g.\@ performing wet-lab experiments for drug design),
making weighted retraining a sensible choice.
\section{Related Work}
\label{sec:related_work}
While a large body of work is applicable to the general problem formulated in \cref{sec:background}
(both with and without machine learning), below we focus only on the most relevant machine learning literature.

\textbf{Latent Space Optimization.}
Early formulations of \gls{lmo} were motivated by scaling Gaussian processes (GPs) to high dimensional problems with simple linear manifolds,
using either random projections \cite{wang_bayesian_2013} or a learned transformation matrix \cite{garnett_active_2013}.
\Gls{lmo} using \glspl{dgm} was first applied to chemical design \cite{gomez2018}, and further built upon subsequently 
\cite{jin_junction_2019,kusner_grammar_2017,eismann_bayesian_2018,kajino_molecular_2019,dai_syntax-directed_2018,daxberger2019bayesian,griffiths_constrained_2020,mahmood_cold_2019}.
It has also been applied to other fields, e.g.\ automatic machine learning \cite{lu2018structured,luo2018neural,zhang_d_vae_2019}, conditional image generation \cite{nguyen_synthesizing_2016,nguyen_plug_2017},
and model explainability \cite{antoran_getting_2020}.
If the surrogate model is a GP, the \gls{dgm} can be viewed as an ``extended kernel'',
making \gls{lmo} conceptually related to deep kernel learning \cite{wilson_deep_2016,huang_scalable_2015}.

\textbf{Weighted Retraining.}
A few previous machine learning methods can be viewed as implementing a version of weighted retraining.
The cross-entropy (CE) method iteratively retrains a generative model using a weighted training set,
such that high-scoring points receive higher weights \cite{rubinstein_optimization_1997,rubinstein_cross-entropy_1999,de2005tutorial}.
Indeed, particular instantiations of the CE method such as reward-weighted regression \cite{peters2007reinforcement},
feedback GAN \cite{gupta_feedback_2019},
and design/conditioning by adaptive sampling (DbAS/CbAS) \cite{Brookes_Listgarten_2020,brookes2019conditioning}
have been applied to similar problem settings as our work.
However, our proposed method of weighted retraining has two main differences from CE.
Firstly, \emph{standard CE produces only binary weights} \cite{de2005tutorial}, which amounts to simply adding or removing points from the training set.%
\footnote{Although methods such as DbAS \cite{Brookes_Listgarten_2020} generalize these weights to lie in $[0,1]$, this is determined by the noise of the oracle and therefore will still produce binary weights when $f$ is deterministic, as considered in this paper.}
This is sub-optimal for reasons discussed in \cref{subsec:weighting,subsec:retraining}, and consequently, \emph{we consider a strictly more general form of weighting}.
Secondly, \emph{CE has no intrinsic optimization component.}
High-performing points are found only by repeatedly sampling from the generative model and evaluating $f$.
By contrast, our method \emph{explicitly selects high-performing points} using Bayesian optimization.
The necessity of repeated sampling in CE makes it only suitable in cases where evaluating $f$ is cheap, which is \emph{not} what we are considering.
Moreover, works such as \cite{segler_generating_2018} perform optimization by fine-tuning a generative model on a smaller dataset of high-scoring samples.
This can also be viewed as a special case of weighted retraining with binary weights,
where the weights are implicitly defined by the number of fine-tuning epochs.

\textbf{\Gls{bo}} is a technique that maintains a probabilistic model of the objective function,
and chooses new points to evaluate based on the modelled distribution of the objective value at unobserved points.
\Gls{bo} is widely viewed as the go-to framework for sample-efficient black-box optimization \cite{brochu2010tutorial,snoek2012practical}.
However, most practical \Gls{bo} models exist for continuous, low-dimensional spaces \cite{shahriari2015taking}.
Recent works have tried to develop models to extend \Gls{bo} to either structured \cite{baptista2018bayesian,kim2019bayesian,daxberger2019mixed,oh2019combinatorial} or high-dimensional \cite{kandasamy2015high,mutny2018efficient,hoang2018decentralized} input spaces.
To our knowledge, only \Gls{bo} methods with a significant amount of domain-specific knowledge infused into their design
are able to handle input spaces that are \emph{both} high-dimensional and structured.
A noteworthy example is ChemBO which uses both a customized molecular kernel and a synthesis graph to perform \Gls{bo} on molecules \cite{korovina2020chembo},
which we compare against in \cref{sec:experiments}.
In contrast, our method can be applied to any problem without domain knowledge, and has comparable performance to ChemBO.
Finally, in an interesting parallel to our work, \cite{blanchard_output_weighted_2020} use a weighted acquisition function to increase the sample efficiency of \gls{bo}.

\textbf{\Gls{rl}} frames optimization problems as Markov decision processes for which an agent learns an optimal policy \cite{sutton1998introduction}.
It has recently been applied to various optimization problems in structured input spaces \cite{li_deep_2018},
notably in chemical design \cite{you_graph_2018,zhou_optimization_2019,guimaraes_objective-reinforced_2018,olivecrona_molecular_2017,popova_deep_2018,simm_reinforcement_2020}.
While \gls{rl} is undoubtedly effective at optimization, it is generally extremely sample inefficient,
and consequently its biggest successes are in virtual environments where function evaluations are inexpensive \cite{li_deep_2018}.

\textbf{Conditional Generative Models.} Finally, one interesting direction is the development of \emph{conditional generative models},
which directly produce novel points conditioned on a specific property value \cite{sohn_learning_2015,mirza_conditional_2014}.
Although many variants of these algorithms have been applied to real-world problems such as chemical design 
\cite{jin_learning_2019,kang_conditional_2019,li_multi-objective_2018,lim_molecular_2018,Brookes_Listgarten_2020},
the sample efficiency of this paradigm is currently unclear.

\section{Empirical Evaluation}
\label{sec:experiments}
This section aims to empirically answer three main questions:
\begin{enumerate}
    \item How does weighted training affect the latent space of \glspl{dgm}? (\cref{subsec:expt-weighting})
    \item How do the parameters of weighted retraining influence optimization? (\cref{subsec:expt-wr})
    \item Does weighted retraining compare favourably to existing methods? (\cref{subsec:expt-baselines})
\end{enumerate}
To answer these questions, we perform experiments using three optimization tasks 
chosen to represent three different data and model types.
The tasks are described in more detail below.
Because there is no obvious single metric to evaluate sample-efficient optimization,
we choose to plot the $K$th best novel evaluated point as a function of the number of objective function evaluations,
which we denote as the \emph{Top$K$} score
(details in \cref{appdx:eval-metric}).
All plots show the average performance and standard deviation across runs with 5 different random seeds unless otherwise stated.
This evaluation method is common practice in Bayesian optimization \cite{shahriari2015taking}.
It contrasts with previous works which typically report only final scores,
and take the maximum across seeds rather than the average
\cite{gomez2018,kusner_grammar_2017,dai_syntax-directed_2018,jin_junction_2019}.

\begin{figure}[t]
    \centering
    \includegraphics{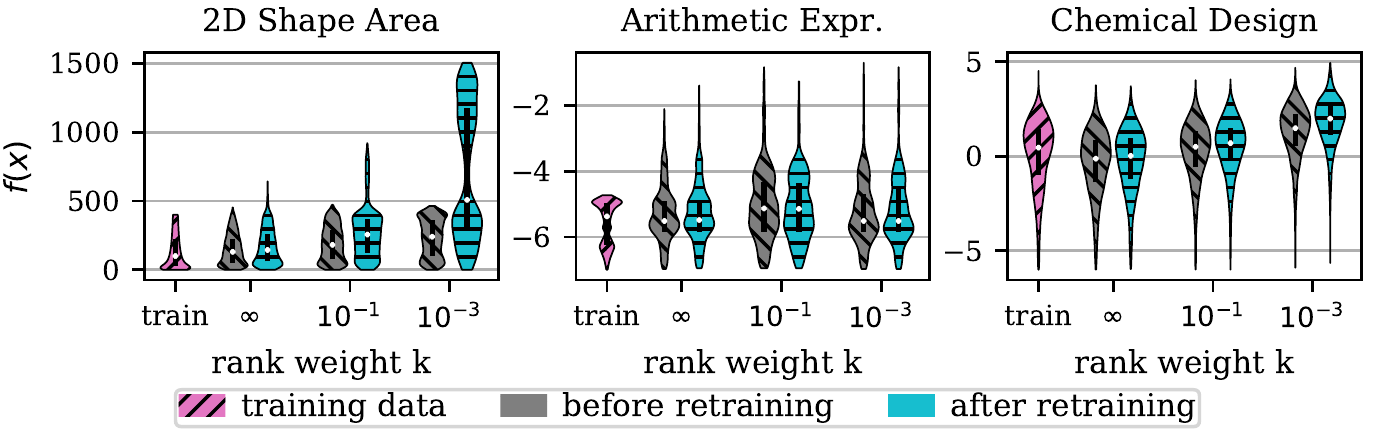}
    \caption{
    Objective value distribution for the training set and samples from the \gls{dgm}'s prior
    for all three tasks for different $k$ values,
    before and after weighted retraining (see \cref{subsec:expt-wr}).
    }
    \label{fig:prior-samples}
\end{figure}

\textbf{2D Shape Area Maximization Toy Task.}
As a simple toy task that can be easily visualized in 2D,
we optimize for the shape with the largest total area in the space of $64\times64$ binary images
(i.e.\@ the largest number of pixels with value 1).
\textbf{Data:} A dataset of $\approx$10,000 squares of different sizes and positions on a $64\times64$ background, with a maximum area of 400 (see \cref{fig:shape-samples} in \cref{sec:exp_details} for examples).
\textbf{Model:} A convolutional \gls{vae} with $\Z=\mathbb{R}^2$, as a standard neural network architecture for image modelling.
\textbf{Latent Optimizer:} We enumerate a grid in latent space over $[-3,+3]^2$, to emulate a perfect optimizer for illustration purposes (this is only feasible since $\Z$ is low-dimensional).

\textbf{Arithmetic Expression Fitting Task.}
We follow \cite{kusner_grammar_2017} and optimize in the space of single-variable arithmetic expressions generated by a formal grammar.
Examples of such expressions are \texttt{sin(2)}, \texttt{v/(3+1)} and \texttt{v/2 * exp(v)/sin(2*v)}, which are all considered to be functions of some variable \texttt{v}.
Following \cite{kusner_grammar_2017}, the objective is to find an expression with minimal mean squared error to the target expression $\x^* = \texttt{1/3 * v * sin(v*v)}$,
computed over 1000 values of \texttt{v} evenly-spaced between $-10$ and $+10$.
\textbf{Data:} 50,000 univariate arithmetic expressions generated by the formal grammar from \cite{kusner_grammar_2017}.
\textbf{Model:} A grammar \gls{vae} \cite{kusner_grammar_2017}, chosen because of its ability to produce only valid grammatical expressions.
\textbf{Latent Optimizer:} Bayesian optimization with the expected improvement acquisition function \cite{jones1998efficient} and a sparse Gaussian process model with 500 inducing points \cite{titsias2009variational},
following \cite{kusner_grammar_2017}.

\textbf{Chemical Design Task.}
We follow \cite{gomez2018} and optimize the drug properties of molecules.
In particular, we consider the standardized task originally proposed in \cite{gomez2018}
of synthesizing a molecule with maximal penalized \emph{water-octanol partition coefficient} (logP),
starting from the molecules in the ZINC250k molecule dataset \cite{irwin_zinc_2012} (see \cref{subsec:appendix-chem-design} for more details).
This task has been studied in a long series of papers performing optimization in chemical space,
allowing the effect of weighted retraining to be quantitatively compared to other optimization approaches
\cite{kusner_grammar_2017,dai_syntax-directed_2018,jin_junction_2019,zhou_optimization_2019,you_graph_2018}.
\textbf{Data:} The ZINC250k molecule dataset \cite{irwin_zinc_2012}, using the same train/test split as \cite{jin_junction_2019}.
\textbf{Model:} A junction tree \gls{vae} \cite{jin_junction_2019},
chosen because it is a state-of-the-art \gls{vae} for producing valid chemical structures.
For direct comparability to previous results,
we use the pre-trained model provided in the code repository of \cite{jin_junction_2019} as the unweighted model,
and create weighted models by fine-tuning the pre-trained model for 1 epoch over the full weighted dataset.
\textbf{Latent Optimizer:} Same as for the arithmetic expression task.

More details on the experimental setup are given in \cref{sec:exp_details}.
All experimental data and code to reproduce the experiments can be found at 
\codelink{}.

\subsection{Effect of Weighted Training}
\label{subsec:expt-weighting}
In this section, we seek to validate some of the conjectures made in \cref{sec:limitations,sec:main},
namely that 1) the latent space of a \gls{dgm} trained on uniformly weighted data contains many poor-performing points, and 2) that weighted training fixes this by introducing more high-performing points into the latent space.
To test this, we train a \gls{vae} for each task using rank weighting with a variety of $k$ values (noting that $k=\infty$ corresponds to uniform weighting), initializing the weights using a pre-trained \gls{vae} to ensure that the different runs are comparable.
We evaluate $f$ on samples from the \gls{dgm}'s prior for each task, and plot the resulting distributions in \cref{fig:prior-samples} with the label \emph{before retraining}.
Although the distribution of scores for $k=\infty$ does not exactly match the training distribution
for any example, it tends to have a similar range, showing that much of the latent space is dedicated to modelling low-scoring points.
Weighted training robustly causes the distribution to skew towards higher values at the expense of lower values, which is exactly the intended effect.
The upshot is that the result on all 3 tasks broadly supports our conjectures.

\begin{figure}[t]
    \centering
    \includegraphics{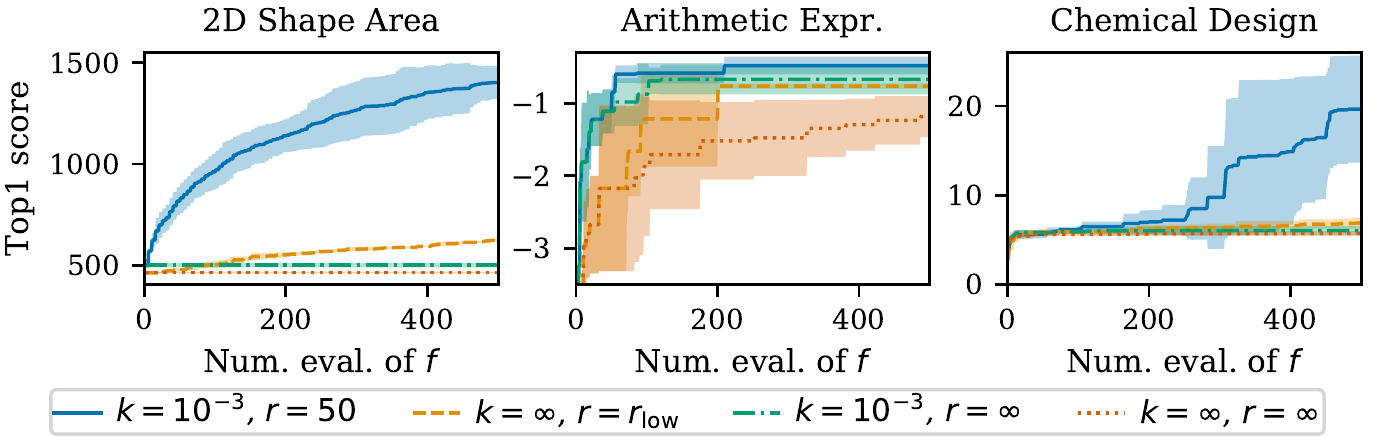}
    \caption{Top$1$ optimization performance of weighted retraining for all tasks, for different $k$ values (i.e. $k \in \{10^{-3}, \infty\}$) and retraining frequencies (i.e. $r_\text{low} = 5$ for the 2D shape area task, and $r_\text{low} = 50$ for the other two tasks). Shaded area corresponds to standard deviation. }
    \label{fig:wr-params}
\end{figure}

\subsection{Effect of Weighted Retraining Parameters on Optimization}
\label{subsec:expt-wr}
When using rank-weighting from \cref{eq:weighting_function} with parameter $k$ and picking a fixed period for model retraining $r$,
\gls{lmo} with weighted retraining can be completely characterized by $k$ and $r$.
The baseline of uniform weighting and no retraining is represented by $k=r=\infty$,
with decreasing values of $k$ and $r$ representing more skewed weighting and more frequent retraining, respectively.
For each task, we choose a value $r_\text{low}$ based on our computational retraining budget,
then perform \gls{lmo} for each value of 
$k\in\{k_{\text{low}}, \infty \}$ and $r\in\{r_{\text{low}}, \infty \}$.
For computational efficiency retraining is done via fine-tuning.
Further experimental details are given in \cref{sec:exp_details}.

The results are shown in \cref{fig:wr-params}.
Firstly, comparing the case of $k=\infty,r=\infty$ with $k=\infty,r=r_{\text{low}}$ and $k=k_{\text{low}},r=\infty$
suggests that both weighting and retraining help individually, as hypothesized in \cref{sec:main}.
Secondly, in all cases, weighted retraining with $k=k_{\text{low}},r=r_{\text{low}}$ performs better than all other methods, suggesting that they have a synergistic effect when combined.
Note that the performance often increases suddenly after retraining,
suggesting that the retraining does indeed incorporate new information into the latent space, as conjectured.
Lastly, the objective function values of prior samples from the models after weighted retraining
with $r=r_{\text{low}}$ is shown in \cref{fig:prior-samples} in blue.
In all cases, the distribution becomes more skewed towards positive values, with the difference being more pronounced for lower $k$ values.
This suggests that weighted retraining is able to significantly modify the latent space, even past the initial retraining.
See \cref{sec:further_results} for results with a larger set of $k$ and $r$ values,
and Top$K$ plots for other values of $K$.

\subsection{Comparison with Other Methods}
\label{subsec:expt-baselines}
\begin{figure}[tb]
    \centering
    \includegraphics[width=\textwidth]{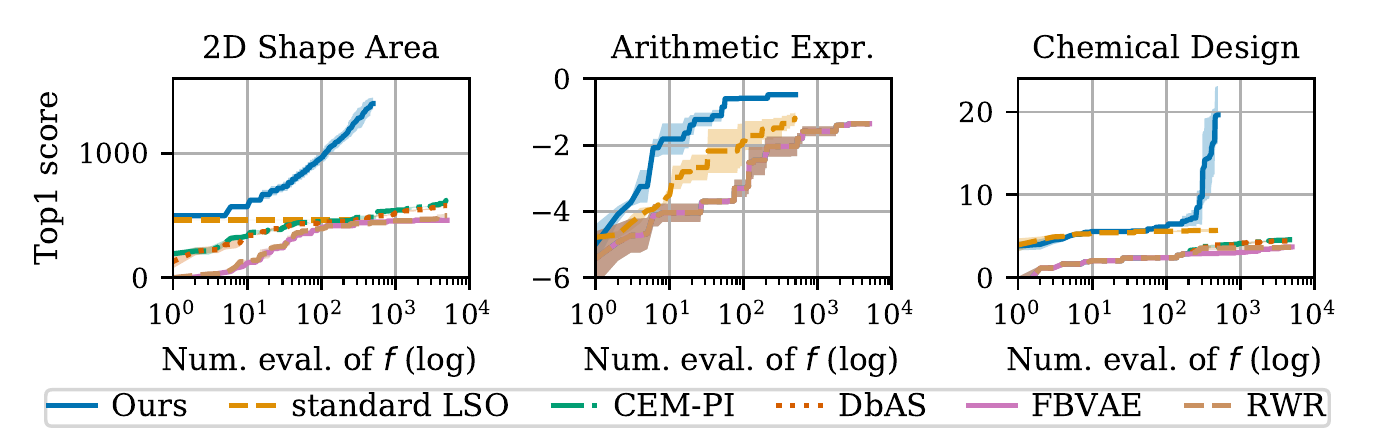}
    \caption{Comparison of weighted retraining, LSO, CEM-PI, DbAS, FBVAE and RWR. Our approach significantly outperforms all baselines, achieving both better sample-efficiency and final performance.}
    \label{fig:baselines}
\end{figure}

Finally, we compare our proposed method of \gls{lmo} with weighted retraining with other methods on the same tasks.
The first class of methods are based on the cross-entropy method as discussed in \cref{sec:related_work},
namely design by adaptive sampling (DbAS) \cite{Brookes_Listgarten_2020}, the cross-entropy method with probability of improvement (CEM-PI) \cite{rubinstein_cross-entropy_1999}, the feedback VAE (FBVAE) \cite{gupta_feedback_2019} and reward-weighted regression (RWR) \cite{peters2007reinforcement}.
These methods are noteworthy because they can be viewed as a particular case of weighted retraining,
where the weights are binary (except for DbAS)
and the latent optimizer simply consists of sampling from the \gls{dgm}'s prior.
The hyperparameters of these methods are the sequence of quantiles, and the retraining frequency.
We optimize these hyperparameters using a grid search, as detailed in \cref{sec:exp_details}.
\Cref{fig:baselines} shows the performance of these methods on the best hyperparameter setting found, as a function of the number of samples drawn (with a budget of 5,000 samples in total).
We plot the average and standard deviation across 3 random seeds, as we found the variances to be relatively low.
We observe that all other forms of weighted retraining perform significantly worse than our own,
failing to achieve the performance of our approach, even with an evaluation budget that is an order of magnitude larger than ours (i.e. 5,000 vs 500).
We attribute this both to their binary weighting scheme and their lack of a sample-efficient latent optimizer.

Secondly, we compare against other methods in the literature that have attempted the same chemical design task.
To our knowledge, the best previously reported score obtained using a machine learning method is 11.84 and was obtained with $\approx5000$ samples \cite{zhou_optimization_2019}.
By contrast, our best score is \bestchemscore{} and was achieved with only 500 samples.
Expanding the scope to include more domain-specific optimization methods,
we acknowledge that ChemBO achieved an impressive score of 18.39 in only 100 samples \cite{korovina2020chembo},
which is better than our method's performance with only 100 samples.
\Cref{tab:chem-results} in the appendix gives a more detailed comparison with other work.
\section{Discussion and Conclusion}
\label{sec:conclusion}
We proposed a method for efficient black-box optimization over high-dimensional, structured input spaces, combining latent space optimization with weighted retraining.
We showed that while being conceptually simple and easy to implement on top of previous methods, weighted retraining significantly boosts their efficiency and performance on challenging real-world optimization problems.

There are several drawbacks to our method that are promising directions for future work.
Firstly, we often found it difficult to train a latent objective model that performed well at optimization in the latent space, which is critical for good performance.
We believe that further research is necessary into other techniques to make the latent space of \glspl{dgm}
more amenable to optimization.
Secondly, our method requires a large dataset of labelled data to train a \gls{dgm},
making it unsuitable for problems with very little data,
motivating adaptations that would allow unlabelled data to be utilized.
Finally, due to being relatively computationally intensive, we were unable to characterize
the long-term optimization behaviour of our algorithm.
In this regime, we suspect that it may be beneficial to use a weighting \emph{schedule}
instead of a fixed weight, which may allow balancing exploration vs.\@ exploitation similar to simulated annealing \cite{van1987simulated}.
Overall, we are excited at the potential results of \gls{lmo} and hope that it can be applied to a variety of real-world problems in the near future.
\section*{Broader Impact}
Ultimately, this work is preliminary, despite the promise of latent space optimization,
there may still be significant obstacles to applying it more widely in the real world.
That aside, we believe that the primary effect of this line of research will be to enable faster discoveries
of novel entities, such as new medicines, new energy materials, or new device designs.
The worldwide effort to develop vaccines and treatments for the COVID-19 pandemic has highlighted
the importance of techniques for fast, targeted discovery using only small amounts of data:
we have seen that even if the whole world is devoted to performing experiments with a single goal,
the sheer size of the search space means that sample efficiency is still important.

As much as this technology could be used to discover good things,
it could also be used to discover bad things (e.g.\@ chemical/biological weapons).
However, as substantial resources and infrastructure are required to produce these,
we do not expect that this line of work will enable new parties to begin their development.
Rather, at worse, it may allow people who are already involved in their development to do it slightly more effectively.

Finally, we believe that this line of work has the potential to influence other problem areas in machine learning,
such as conditional image generation and conditional text generation,
because these tasks can also be viewed as optimization, whose objective function is a human judgement,
which is generally expensive to obtain.

\begin{ack}
We thank Ross Clarke, Gregor Simm, David Burt,
and Javier Antor\'{a}n for insightful feedback and discussions.
AT acknowledges funding via a C T Taylor Cambridge International Scholarship.
ED acknowledges funding from the EPSRC and Qualcomm.
This work has been performed using resources provided by the Cambridge Tier-2 system operated by the University of Cambridge Research Computing Service (http://www.hpc.cam.ac.uk) funded by EPSRC Tier-2 capital grant EP/P020259/1.
\end{ack}

\bibliographystyle{abbrv}
\bibliography{neurips_2020}

\newpage
\appendix
\section{Details on the Weighting Function}
\label{sec:weighting-details}
\subsection{More Information on Rank-Based Weighting}
\label{subsec:rank-weights}
\paragraph{Independence from Dataset Size}
We show that the key properties of rank-based weighting depend \emph{only} on $k$,
and not on the dataset size $N$, meaning that applying rank weighting with a fixed $k$ to differently sized datasets will yield similar results.
In particular, we show that under mild assumptions, the fraction of weights devoted to a particular quantile of the data depends on $k$ but not $N$.

Suppose that the quantile of interest is the range $q_1$--$q_2$ (for example, the first quartile corresponds to the range $0$--$0.25$).
This corresponds approximately to the points with ranks $q_1N$--$q_2N$.
We make the following assumptions:
\begin{enumerate}
    \item $kN\gg 1$
    \item $kN$ is approximately integer valued, which is realistic if $N\gg 1/k$
    \item $q_1$ and $q_2$ are chosen so that $q_1N$ and $q_2N$ are integers.
\end{enumerate}

\begin{figure}
    \centering
    \includegraphics[width=0.49\linewidth]{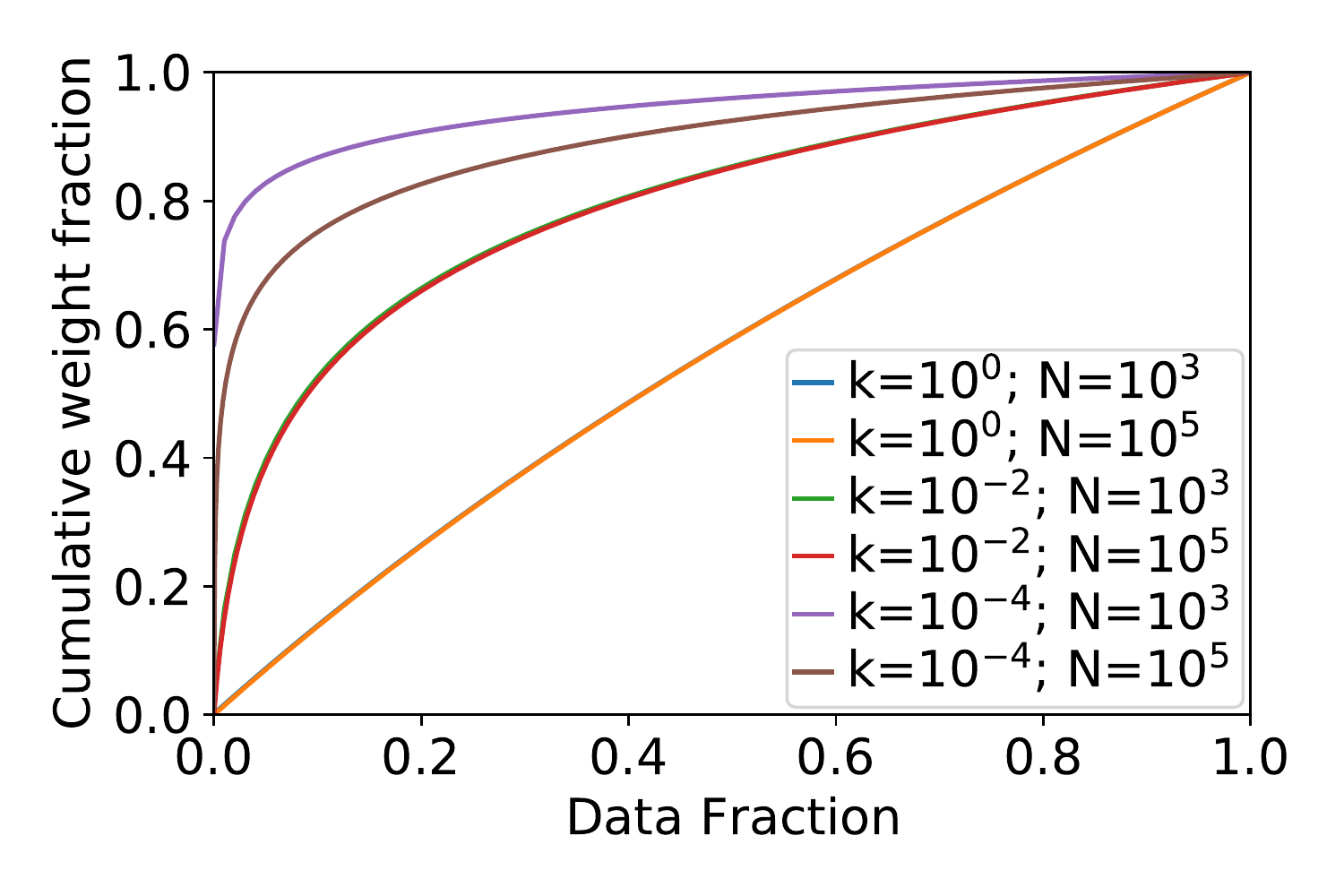}
    \caption{Cumulative distribution of rank weights (sorted highest to lowest), showing a distribution that is independent of $N$ if $kN>1$.}
    \label{fig:weight-sum-dists}
\end{figure}

Because the ranks form the sequence $0,1,\ldots,N-1$,
under the above assumptions all weights are reciprocal integers,
so the sum of the rank weights is strongly connected to the harmonic series.
Recall that the partial sum of the harmonic series can be approximated by the natural logarithm:
\begin{equation}
    \label{eqn:harmonic-series}
    \sum_{j=1}^N \frac{1}{j} \approx \ln{N} + \gamma
\end{equation}
Here, $\gamma$ is the Euler–Mascheroni constant.
The fraction of the total weight devoted to the quantile $q_1$--$q_2$ can be found by summing the weights  of points with rank $q_1N$--$q_2N$, and dividing by the normalization constant (the sum of all weights). 
First, because $kN\gg1$ implies that $(kN-1)\approx kN$, the sum of all the weights can be expressed as:
\begin{align*}
    \sum_{r=0}^{N-1}w(\x_r; \D, k) &= \sum_{r=0}^{N-1} \frac{1}{kN + r} \\
     &= \sum_{r=1}^{kN+(N-1)} \frac{1}{r} - \sum_{r'=1}^{kN-1} \frac{1}{r'} \\
     &\approx \left(\ln{\left((k+1)N-1\right)} + \gamma\right) - \left(\ln{(kN-1)} + \gamma\right) \\
     &=  \ln{\frac{(k+1)N-1}{kN-1}} \approx \ln{\frac{(k+1)N}{kN}}= \ln{\left(1 + \frac{1}{k}\right)} \\
\end{align*}
Note that this does not depend on the dataset size $N$.
Second, using the same assumption, the sum of the weights in the quantile is:
\begin{align*}
    \sum_{r=q_1N}^{q_2N}w(\x_r; \D, k) &= \sum_{r=q_1N}^{q_2N} \frac{1}{kN + r} \\
     &= \sum_{r=1}^{(k+q_2)N} \frac{1}{r} - \sum_{r'=1}^{(k+q_1)N-1} \frac{1}{r'} \\
     &\approx \left(\ln{\left((k+q_2)N-1\right)} + \gamma\right) - \left(\ln{((k+q_1)N-1)} + \gamma\right) \\
     &=  \ln{\frac{(k+q_2)N}{(k+q_1)N-1}} \approx \ln{\frac{(k+q_2)N}{(k+q_1)N}} = \ln{\frac{(k+q_2)}{(k+q_1)}} \\
\end{align*}
which is also independent of $N$
(note that setting $q_1=0$, $q_2=1$ into the formula yields the same expression for the sum of the weights as derived above).
Therefore, the fraction of the total weight allocated to a given quantile of data is \emph{independent of $N$},
being only dependent on $k$.
Although the analysis that led to this result made some assumptions about certain values being integers,
in practice the actual distributions of weights are extremely close to what this analysis predicts.
\cref{fig:weight-sum-dists} shows the allocation of the weights to different quantiles of the datasets.
For $kN>1$, the distribution is essentially completely independent of $N$.
Only when $kN<1$ this fails to hold.

Finally, we discuss some potential questions about the rank-based weighting.

\paragraph{Why do the weights need to be normalized?}
If the objective is to minimize $\sum_{\x_i\in\D}w_i\mathcal{L}(\x_i)$,
for any $a>0$, minimizing $a\sum_{\x_i\in\D}w_i\mathcal{L}(\x_i)$ is an equivalent problem.
Therefore, in principle, the absolute scale of the weights does not matter,
and so the weights do not \emph{need} to be normalized, even if this precludes their interpretation as a probability distribution.
However, in practice, if minimization is performed using gradient-based algorithms,
then the scaling factor for the weights is also applied to the gradients,
possibly requiring different hyperparameter settings (such as a different learning rate).
By normalizing the weights, it is easier to identify hyperparameter settings that work robustly across different problems,
thereby allowing weighted retraining to be applied with minimal tuning.

\paragraph{Why not use a weight function directly based on the objective function value?}
Although there is nothing inherently flawed about using such a weight function,
there are some practical difficulties.
\begin{itemize}
    \item Such a weight function would either be bounded (in which case values beyond a certain threshold would all be weighted equally),
    or it would be very sensitive to outliers (i.e.\@ extremely high or low values which would directly cause the weight function to take on an extremely high or low value).
    This is extremely important because the weights are \emph{normalized},
    so one outlier would also affect the values of all other points.
    \item Such a weight function would not be invariant to simple transformations of the objective function.
    For example, if the objective function is $f$, then maximizing $f(\x)$ or $f_{ab}(\x) = af(\x)+b$ is an equivalent problem (for $a>0$),
    but would yield different weights.
    This would effectively introduce scale hyperparameters into the weight function,
    which is undesirable.
\end{itemize}

\subsection{Mini-Batching for Weighted Training}
\label{subsec:weighted-mini-batching}

As mentioned in the main text, one method of implementing the weighting with mini-batch stochastic gradient descent
is to sample each point $x_i$ with probability proportional to its weight $w_i$ (with replacement).
A second method is to sample points with uniform probability and re-weight each point's contribution to the total loss
by its weight:
\begin{equation}
\label{eq:dgm_loss_minibatched}
    \sum_{\x_i \in \D} w_i \mathcal{L}(\x_i) \approx \frac{N}{n}\sum_{j=1}^n w_j \mathcal{L}(\x_j)
\end{equation}
If done naively, these mini-batches may have extremely high variance, especially if the variance of the weights is large.
In practice, we found it was sufficient to reduce the variance of the weights by simply adding multiple copies of any $\x_i$
with $w_i>w_{\text{max}}$, then reducing the weight of each copy such that the sum is still $w_i$.
The following is a \texttt{Python} code snippet implementing this variance reduction:
\begin{minted}{Python}
def reduce_variance(data, weights, w_max):
    new_data = []
    new_weights = []
    for x, w in zip(data, weights):
        if w <= w_max:  # If it is less than the max weight, just add it
            new_data.append(x)
            new_weights.append(w)
        else:  # Otherwise, add multiple copies
            n_copies = int(math.ceil(w / w_max))
            new_data += [x] * n_copies
            new_weights += [w / n_copies] * n_copies
    return new_data, new_weights
\end{minted}
The parameter \texttt{w\_max} was typically set to $5.0$, which was chosen to both reduce the variance,
while simultaneously not increasing the dataset size too much.
Note that this was applied \emph{after} the weights were normalized.
This also makes it feasible to train for only a fraction of an epoch,
since without variance reduction techniques there is a strong possibility that high-weight data points
would be missed if the entire training epoch was not completed.

\subsection{Implementation of Weighted Training}
\label{subsec:weighted-training-implementation}

One of the benefits of weighted retraining which we would like to highlight is its ease of implementation.
Below, we give example implementations using common machine learning libraries.

\subsubsection{PyTorch (weighted sampling)}

\begin{minipage}{.45\linewidth}
\textbf{Standard Training}
\begin{minted}{Python}
from torch.utils.data import *


dataloader = DataLoader(data)
for batch in dataloader:
    # ...
\end{minted}
\end{minipage}\hfill
\begin{minipage}{.45\linewidth}
\textbf{Weighted Training}
\begin{minted}{Python}
from torch.utils.data import *
sampler = WeightedRandomSampler(
    weights, len(data))
dataloader = DataLoader(data, sampler=sampler)
for batch in dataloader:
    # ...
\end{minted}
\end{minipage}

\subsubsection{PyTorch (direct application of weights)}

\begin{minipage}{.45\linewidth}
\textbf{Standard Training}
\begin{minted}{Python}
criterion = nn.MSELoss()
outputs = model(inputs)
loss = criterion(outputs, targets)

loss.backward()
\end{minted}
\end{minipage}\hfill
\begin{minipage}{.45\linewidth}
\textbf{Weighted Training}
\begin{minted}{Python}
criterion = nn.MSELoss(reduction=None)
outputs = model(inputs)
loss = criterion(outputs, targets)
loss = torch.mean(loss * weights)
loss.backward()
\end{minted}
\end{minipage}

\subsubsection{Keras}

\begin{minipage}{.45\linewidth}
\textbf{Standard Training}
\begin{minted}{Python}
model.fit(x, y)
\end{minted}
\end{minipage}\hfill
\begin{minipage}{.45\linewidth}
\textbf{Weighted Training}
\begin{minted}{Python}
model.fit(x, y, sample_weight=weights)
\end{minted}
\end{minipage}

\subsection{Implementation of Rank Weighting}
We provide a simple implementation of rank-weighting:

\begin{minted}{Python}
import numpy as np
def get_rank_weights(outputs, k):

    # argsort argsort to get ranks (a cool trick!)
    # assume here higher outputs are better
    outputs_argsort = np.argsort(-np.asarray(outputs))
    ranks = np.argsort(outputs_argsort)
    return 1 / (k * len(outputs) + ranks)
\end{minted}

\subsection{Rank-Weighted Distributions of Objective Function Values of 2D Shape and Arithmetic Expression Datasets}
\begin{figure}[ht]
    \hspace{-4mm}
    \includegraphics{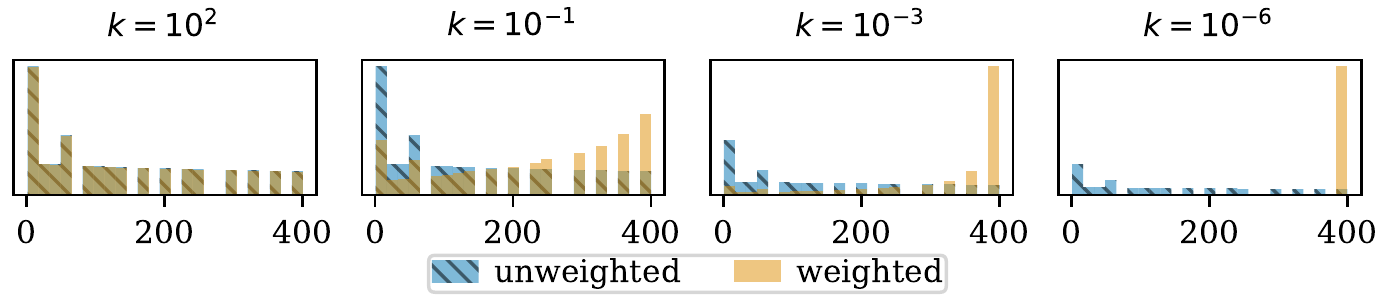}
    \hspace{-4.5mm}
    \caption{
    Illustration of rank weighting (\cref{eq:weighting_function})
    on the shapes dataset (see \cref{sec:experiments}) (similar to \cref{fig:weighting}).
    }
    \label{fig:weighting_shape}
\end{figure}

\begin{figure}[ht]
    \hspace{-4mm}
    \includegraphics{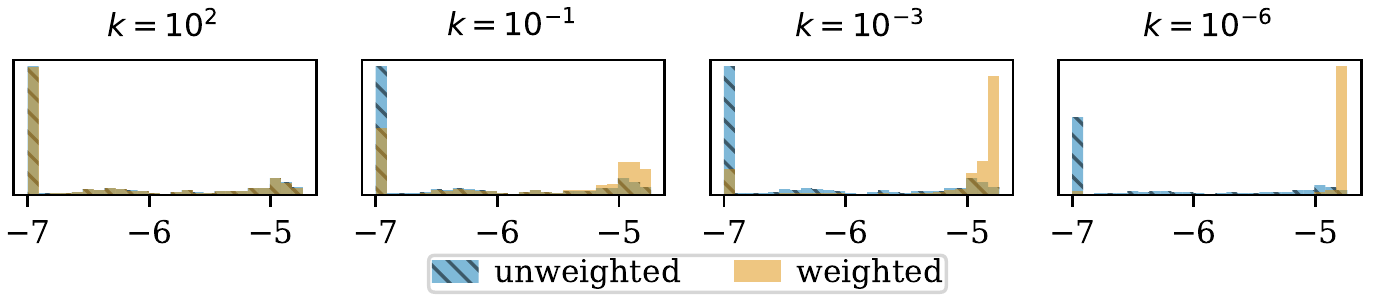}
    \hspace{-4.5mm}
    \caption{
    Illustration of rank weighting (\cref{eq:weighting_function})
    on the arithmetic expression dataset (see \cref{sec:experiments}) (similar to \cref{fig:weighting}).
    }
    \label{fig:weighting_equation}
\end{figure}

Finally, to complement the rank-weighted distributions of objective function values of the ZINC dataset in \cref{fig:weighting}, we here also show the corresponding distributions for the 2D shape and arithmetic expression datasets used in \cref{sec:experiments}
(\cref{fig:weighting_shape} and \cref{fig:weighting_equation}).
\section{Further Experimental Results}
\label{sec:further_results}
\subsection{Optimization Performance with More Weighted Retraining Parameters}

\begin{figure}[tb]
    \centering
    \includegraphics{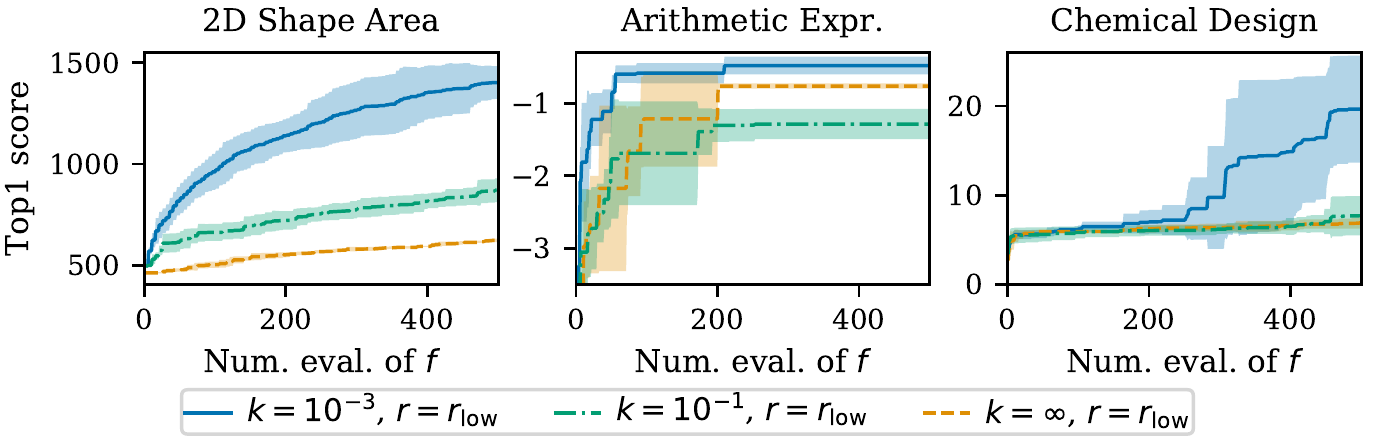}
    \caption{
        Top1 optimization performance of weighted retraining for different $k$ values with $r=r_{\text{low}}$.
    }
    \label{fig:vary_k}
\end{figure}

\begin{figure}[tb]
    \centering
    \includegraphics{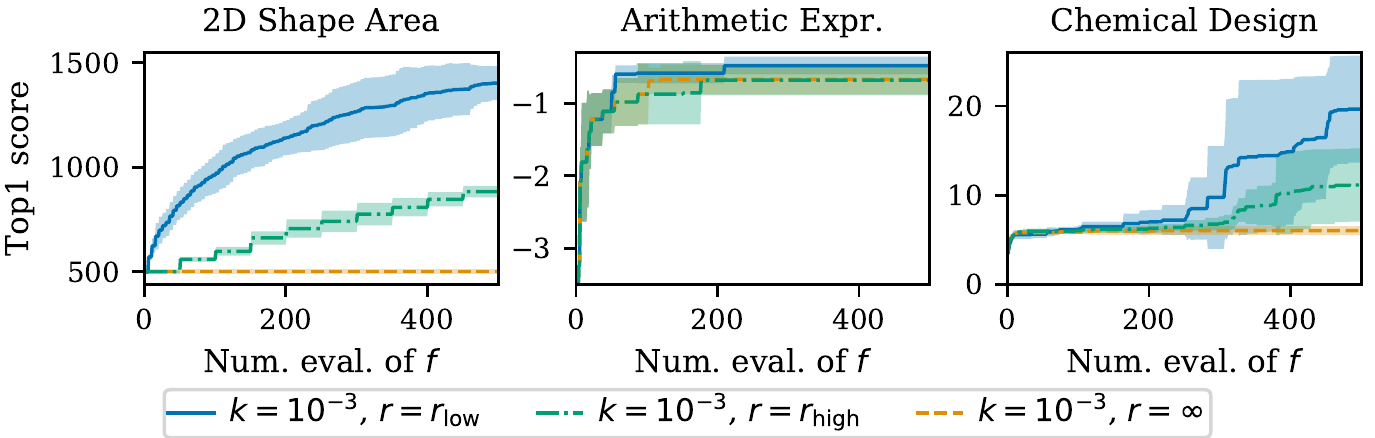}
    \caption{
        Top1 optimization performance of weighted retraining for different $r$ values with $k=k_{\text{low}}$
        $r_{\text{low}}=5,r_{\text{high}}=50$ for the 2D shape area task; 
        $r_{\text{low}}=50,r_{\text{high}}=100$ for the others
    }
    \label{fig:vary_r}
\end{figure}

Holding $r$ fixed at $r_\text{low}$, we vary $k$ from $k_\text{low}$ to $\infty$
(\cref{fig:vary_k}) and vice versa (\cref{fig:vary_r}).
In general, performance increases monotonically as $k,r$ decrease,
suggesting a continuous improvement from increasing weighting or retraining.
The arithmetic expression task did not show this behaviour for retraining,
which we attribute to the high degree of randomness in the optimization.


\subsection{Top10 and Top50 Optimization Results}

\begin{figure}[tb]
    \centering
    \includegraphics{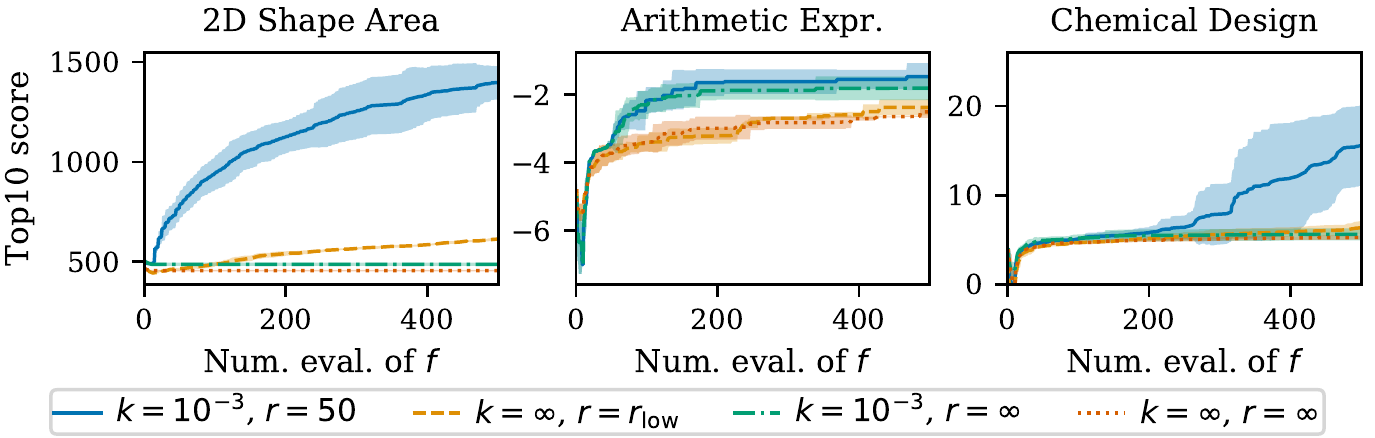}
    \caption{
        Top10 optimization performance of weighted retraining for all tasks (setup identical to \cref{fig:wr-params}).
    }
    \label{fig:wr_params_top10}
\end{figure}

\begin{figure}[tb]
    \centering
    \includegraphics{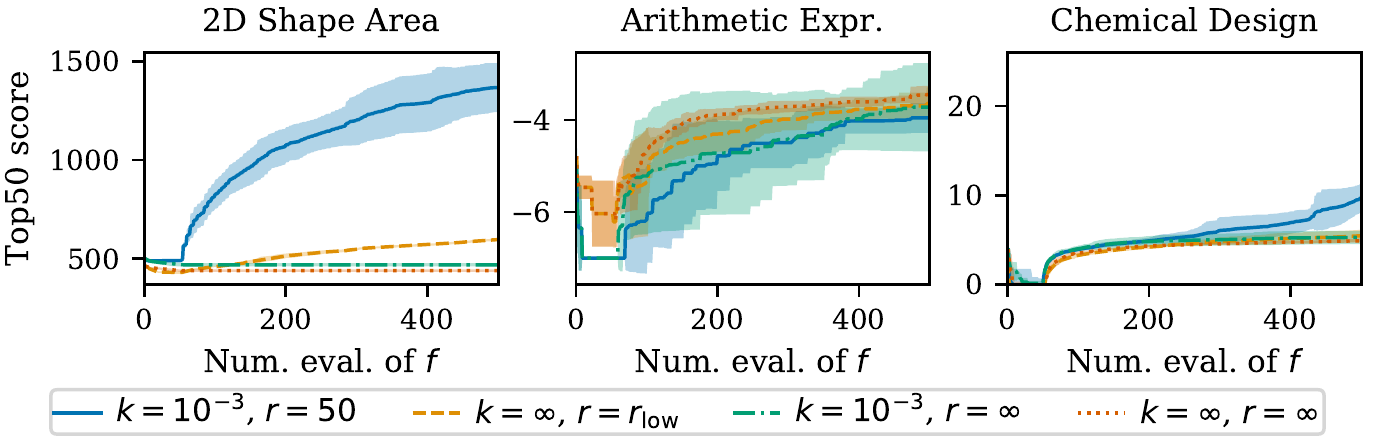}
    \caption{
        Top50 optimization performance of weighted retraining for all tasks (setup identical to \cref{fig:wr-params}).
    }
    \label{fig:wr_params_top50}
\end{figure}

\Cref{fig:wr_params_top10} and \cref{fig:wr_params_top50}
give the Top10 and Top50 scores for the experiment described in \cref{subsec:expt-wr}.
These results are qualitatively similar to those in \cref{fig:wr-params},
suggesting that our method finds many unique high-scoring points.



\clearpage
\subsection{Comparison of Chemical Design Results with Previous Papers}
\Cref{tab:chem-results} compares the results attained in this paper with the results from previous papers that attempted the same task.
Weighted retraining clearly beats the previous best methods, which were based on reinforcement learning,
while simultaneously being more sample-efficient.
Note that despite using the same pre-trained model as \cite{jin_junction_2019},
we achieved better results by training our sparse Gaussian process on only a subset of data
and clipping excessively low values in the training set,
which allowed us to get significantly better results than they reported.
\begin{table}[h]
  \centering
  \begin{tabular}{lllll}
    \toprule
    \textbf{Model}     & \textbf{1st}     & \textbf{2nd} & \textbf{3rd} & \textbf{no.\@ queries (source)} \\
    \midrule
    JT-VAE \cite{jin_junction_2019} & 5.30  & 4.93 & 4.49 & 2500 (paper\tablefootnote{These were the top results across 10 seeds, with 250 queries performed per seed.})     \\
    GCPN \cite{you_graph_2018}     & 7.98 & 7.85 & 7.80 & $\approx10^6$ (email\tablefootnote{Obtained through email correspondence with the authors.})     \\
    MolDQN \cite{zhou_optimization_2019} & 11.84 & 11.84 & 11.82 & $\geq5000$ (paper\tablefootnote{The experimental section states that the model was trained for 5000 episodes, so at least 5000 samples were needed. It is unclear if any batching was used, which would make the number of samples greater.})  \\
    ChemBO \cite{korovina2020chembo} & 18.39 & - & - & \textbf{100} (Table 3 of \cite{korovina2020chembo}) \\
    \midrule
    \textbf{JT-VAE} (our Bayesian optimization) & 5.65 & 5.63 & 5.43 & 500 \\
    \textbf{JT-VAE} ($k=10^{-3}$, no retraining) & 5.95 & 5.75 & 5.72 & 500 \\
    \textbf{JT-VAE} ($k=10^{-3}$, retraining) & 21.20 & 15.34 & 15.34 & 500 \\
    \textbf{JT-VAE} ($k=10^{-3}$, retraining, \emph{best result}) & \textbf{27.84} & \textbf{27.59} & \textbf{27.21} & \textbf{500} \\
    \bottomrule
  \end{tabular}
  \caption{Comparison of top 3 scores on chemical design task.
  Baseline results are copied from \cite{zhou_optimization_2019}.
  All our results state the \emph{median} of 5 runs unless otherwise stated (judged by best result),
  each run being 500 epochs.
  }
  \label{tab:chem-results}
\end{table}

\subsection{Pictures of the Best Molecules Found by Weighted Retraining}
\begin{figure}[ht]
    \centering
    \begin{tikzpicture}
        \draw (0, 0) node[inner sep=0] {\includegraphics[width=0.3\linewidth]{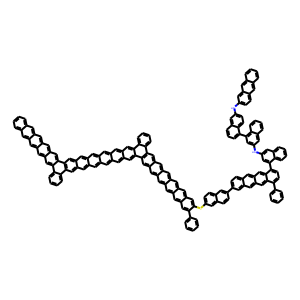}};
        \draw (0, 1) node {27.84};
    \end{tikzpicture}
    \begin{tikzpicture}
        \draw (0, 0) node[inner sep=0] {\includegraphics[width=0.3\linewidth]{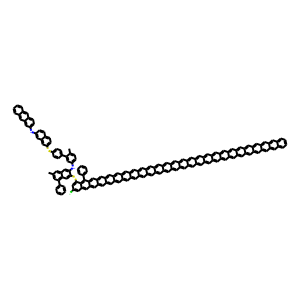}};
        \draw (1, 1) node {27.59};
    \end{tikzpicture}
    \begin{tikzpicture}
        \draw (0, 0) node[inner sep=0] {\includegraphics[width=0.3\linewidth]{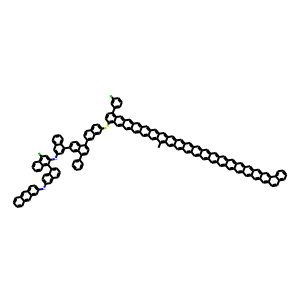}};
        \draw (1, 1) node {27.21};
    \end{tikzpicture}
    \begin{tikzpicture}
        \draw (0, 0) node[inner sep=0] {\includegraphics[width=0.3\linewidth]{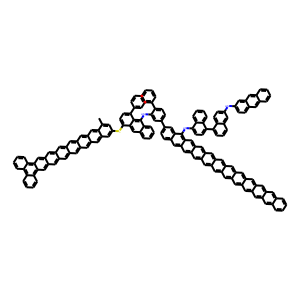}};
        \draw (1, 1) node {25.90};
    \end{tikzpicture}
    \begin{tikzpicture}
        \draw (0, 0) node[inner sep=0] {\includegraphics[width=0.3\linewidth]{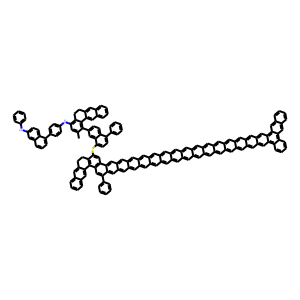}};
        \draw (1, 1) node {25.37};
    \end{tikzpicture}

    \caption{Some of the best molecules found using weighted retraining. Numbers indicate the score of each molecule.}
    \label{fig:best-molecule-pix}
\end{figure}
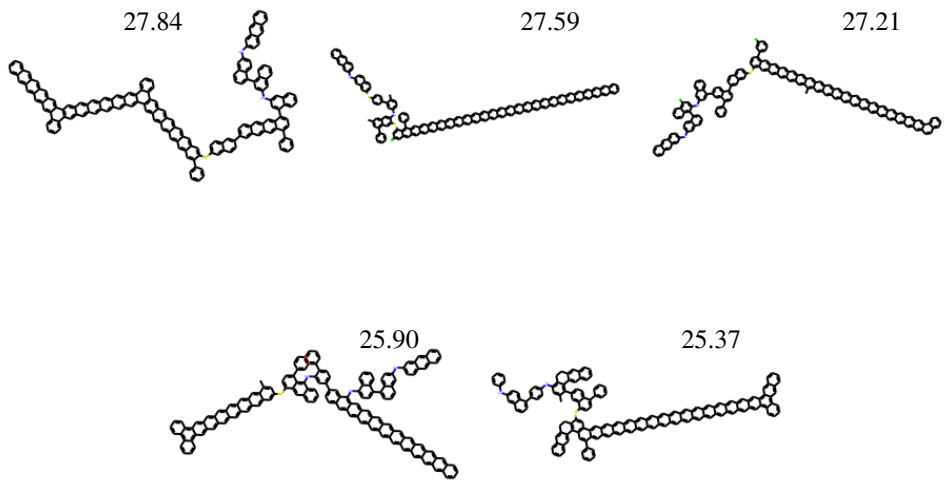

\Cref{fig:best-molecule-pix} illustrates some of the best molecules found with weighted retraining.
Note that all the high-scoring molecules are extremely large.
It has been reported previously that larger molecules achieve higher scores,
thereby diminishing the value of this particular design task for \gls{rl} algorithms \cite{zhou_optimization_2019}.
However, the fact that these molecules were found with a generative model strongly highlights
the ability of weighted retraining to find solutions outside of the original training distribution.

\section{Details on Experimental Setup}
\label{sec:exp_details}
\subsection{Retraining Parameters}
When retraining a model with frequency $r$,
the model is optionally fine-tuned initially, then repeatedly fine-tuned on queries $r,\ 2r,\ 3r,\ldots$
until the query budget is reached.
All results use the rank-based weighting function defined in \cref{eq:weighting_function}
unless otherwise specified.
We consider a budget of $B=500$ function evaluations, which is double the budget used in \cite{kusner_grammar_2017,jin_junction_2019}.

\subsection{Bayesian Optimization}
For optimizing over the latent manifold, we follow previous work \cite{kusner_grammar_2017,jin_junction_2019} and use Bayesian optimization with a variational \glsfirst{sgp} surrogate model \cite{titsias2009variational} (with 500 inducing points) and the expected improvement acquisition function \cite{jones1998efficient}.
We re-implemented the outdated and inefficient \texttt{Theano}-based Bayesian optimization implementation of \cite{kusner_grammar_2017} (see \url{https://github.com/mkusner/grammarVAE}), which was also used by \cite{jin_junction_2019}, using the popular and modern \texttt{Tensorflow 2.0}-based \texttt{GPflow} Gaussian process library \cite{de2017gpflow} to benefit from GPU acceleration.

For computational efficiency, we fit the \gls{sgp} only on a subset of the data, consisting of the 2000 points
with the highest objective function values,
and 8000 randomly chosen points.
This also has the effect of ensuring that the \gls{sgp} properly fits the high-performing regions of the data.
Disregarding computational efficiency, we nonetheless found that fitting on this data subset remarkably improved performance
of the optimization, even using the baseline model (without weighted retraining).

\subsection{Evaluation Metrics}
\label{appdx:eval-metric}
We report, as a function of the objective function evaluation $b=1,\ldots,B$, the single best score obtained up until query $b$ (denoted as Top1), and the worst of the 10 and 50 best scores obtained up until evaluation query $b$ (denoted as Top10 and Top50, respectively).
Since our goal is to synthesize entities with the desired properties that are both a) \emph{syntactically valid} and b) \emph{novel}, we discard any suggested data points which are either a) invalid or b) contained in the training data set (i.e., they are not counted towards the evaluation budget and thus not shown in any of the plots).
For statistical significance, we always report the mean plus/minus one standard deviation across multiple random seeds.

\subsection{2D Shape Task Details}
\label{subsec:appendix-shapes}
\begin{figure}[ht]
    \centering
    \includegraphics[width=\textwidth]{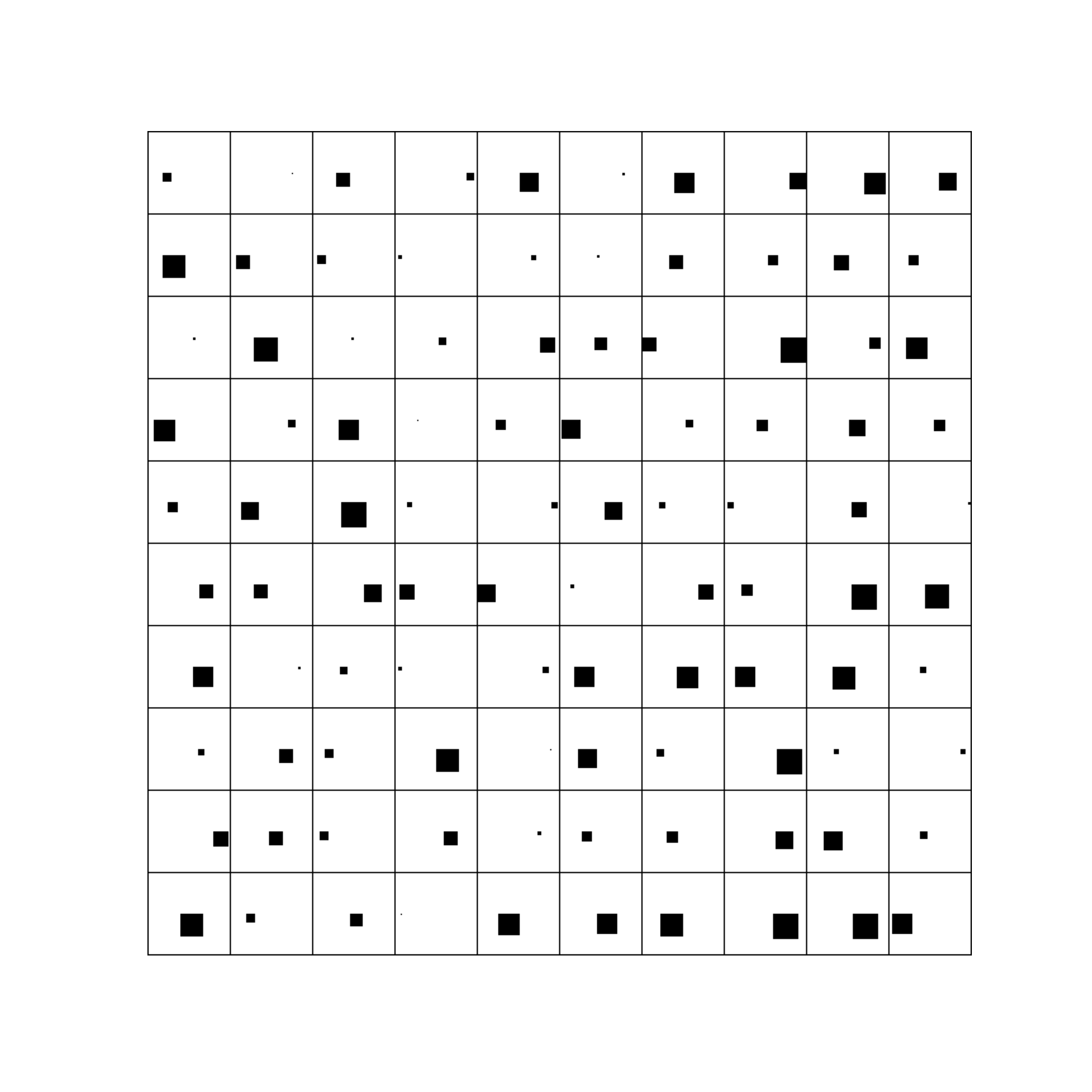}
    \caption{Sample images from our 2D squares dataset.}
    \label{fig:shape-samples}
\end{figure}

\cref{fig:shape-samples} shows example images from our 2D squares dataset.

The convolutional \gls{vae} architecture may be found in our code.
The decoder used an approximately mirror architecture to the encoder with transposed convolutions.
Following general conventions, we use a standard normal prior $p(\z) = \mathcal{N}(0,1)$ over the latent variables $\z$ and a Bernoulli likelihood $p(\x|\z)$ to sample binary images.
Our implementation used PyTorch \cite{paszke2017automatic} and PyTorch Lightning \cite{falcon2019pytorch}.

\subsection{Arithmetic Expression Fitting Task}
\label{subsec:appendix-equation}
Following \cite{kusner_grammar_2017}, the dataset we use consists of randomly generated univariate arithmetic expressions from the following grammar:
\begin{align*}
    &\texttt{S $\rightarrow$ S '+' T | S '*' T | S '/' T | T}\\
    &\texttt{T $\rightarrow$ '(' S ')' | 'sin(' S ')' | 'exp(' S ')'}\\
    &\texttt{T $\rightarrow$ 'v' | '1' | '2' | '3'}
\end{align*}
where \texttt{S} and \texttt{T} denote non-terminals and the symbol \texttt{|} separates the possible production rules generated from each non-terminal.
Every string in the dataset was generated by applying at most 15 production rules, yielding arithmetic expressions such as \texttt{sin(2)}, \texttt{v/(3+1)} and \texttt{v/2 * exp(v)/sin(2*v)}, which are all considered to be functions of the variable \texttt{v}.

The objective function we use is defined as $f(\x) = -\log(1 + \text{MSE}(\x, \x^*))$, where $\text{MSE}(\x, \x^*)$ denotes the mean squared error between $\x$ and the target expression $\x^* = \texttt{1/3 * v * sin(v*v)}$,
computed over 1000 evenly-spaced values of \texttt{v} in the interval between $-10$ and $+10$.
We apply the logarithm function following \cite{kusner_grammar_2017} to avoid extremely large MSE values resulting from exponential functions in the generated arithmetic expressions.
In contrast to \cite{kusner_grammar_2017}, we negate the logarithm to arrive at a maximization problem (instead of a minimization problem), to be consistent with our problem formulation and the other experiments.
The global maximum of this objective function is $f(\x) = 0$, achieved at $\x = \x^*$ (and $f(\x) < 0$ otherwise).

In contrast to the original dataset of size 100,000 used by \cite{kusner_grammar_2017}, which \emph{includes the target expression} and many other well-performing inputs (thus making the optimization problem easy in theory), we make the task more challenging by discarding the 50\% of points with the highest scores, resulting in a dataset of size 50,000 with objective function value distribution shown in \cref{fig:weighting_equation}.

Our implementation of the grammar \gls{vae} is based on the code from 
\cite{kusner_grammar_2017} provided at \url{https://github.com/mkusner/grammarVAE},
which we modified to use Tensorflow 2 \cite{abadi2016tensorflow} and python 3.

\subsection{Chemical Design Task}
\label{subsec:appendix-chem-design}
The precise scoring function for a chemical $\x$ is defined as:
\begin{equation*}
\text{score}(\x) = \max\left(\widehat{\log{P}(\x)} -\widehat{\text{SA}(\x)} - \widehat{\text{cycle}(\x)}, \ -4\right)
\end{equation*}
where $\log{P}$, SA, and cycle are property functions,
and the $\ \widehat{}\ $ operation indicates standard normalization of the raw function output using the ZINC training set data
(i.e.\@ subtracting the mean of the training set, and dividing by the standard deviation).
This is identical to the scoring function from references \cite{kusner_grammar_2017,dai_syntax-directed_2018,jin_junction_2019,zhou_optimization_2019,you_graph_2018},
except that we bound the score below by $-4$ to prevent points with highly-negative scores from substantially impacting the optimization procedure.
Functionally, because this is a maximization task, this makes little difference to the scoring of the outcomes,
but does substantially help the optimization.

Our code for the junction tree \gls{vae} is a modified version of the ``fast jtnn'' code from
the authors of  \cite{jin_junction_2019} (available at \url{https://github.com/wengong-jin/icml18-jtnn}).
We adapted the code to be backward-compatible with their original pre-trained model,
and to use pytorch lightning.

\subsection{Other Reproducibility Details}

\paragraph{Range of hyperparameters considered}
We originally considered $k$ values in the range $10^1, 10^0, \ldots, 10^{-5}$, and found that there was generally a regime where improvement was minimal,
but below a certain $k$ value there was significant improvement (which is consistent with our theory).
We chose $k=10^{-3}$ as an intermediate value that consistently gave good performance across tasks.
This value was chosen in advance of running our final experiments (i.e.\@ we had preliminary but incomplete results with other $k$ values, then chose $k=10^{-3}$,
and then got our main results).
The retraining frequency of 50 was chosen arbitrarily in advance of doing the experiments
(specifically it was chosen because it would entail retraining 10 times in our 500 epochs of optimization).
The hyperparameters for model design and learning were dictated by the papers whose models we chose,
except for the convolutional neural network for the shape task,
where we chose a generic architecture.
For the baseline methods (i.e.\ DbAS, CEM-PI, FBVAE, and RWR), we identified the best hyperparameter settings using a grid search over a reasonable range.
We used the following hyperparameter settings: a quantile parameter of 0.95 for DbAS, CEM-PI and FBVAE (for all benchmarks), a retrain frequency of 200 for all baselines and for all benchmarks, an exponential coefficient of $10^{-3}$ (for the shapes task) and $10^{-1}$ (for the expression and chemical design tasks) for RWR, and a noise variance of 10 (for the shapes task) and 0.1 (for the expression and chemical design tasks) for DbAS.

\paragraph{Average run time for each result}
All experiments were performed using a single GPU.
Runtime results are given in \cref{tab:runtimes}.

\begin{table}[h]
  \centering
  \begin{tabular}{ll}
    \toprule
    \textbf{Experiment}     & \textbf{GPU hours per run} \\
    \midrule
    Shapes (model pre-training) & 0:20  \\
    Shapes (optim., retraining) & 0:20  \\
    Shapes (optim., no retraining) & 0:01  \\
    \midrule
    Expressions (optim., retraining) & 3:15  \\
    Expressions (optim., no retraining) & 1:45  \\
    \midrule
    Chemical Design (optim., retraining) & 5:00  \\
    Chemical Design (optim., no retraining) & 3:00  \\
    \bottomrule
  \end{tabular}
  \caption{Approximate runtimes of main experiments
  }
  \label{tab:runtimes}
\end{table}

\paragraph{Computing infrastructure used}
All experiments were done using a single GPU (either NVIDIA P100, 2070 Ti, or 1080 Ti).
In practice, a lot of the experiments were run on a high-performance computing cluster to allow multiple experiments to be run in parallel,
although this was strictly for convenience: in principle, all experiments could be done on a single machine with one GPU.

\end{document}